\documentclass[a4paper,fleqn]{cas-dc}

\usepackage[numbers]{natbib}

\def\tsc#1{\csdef{#1}{\textsc{\lowercase{#1}}\xspace}}
\tsc{WGM}
\tsc{QE}
\tsc{EP}
\tsc{PMS}
\tsc{BEC}
\tsc{DE}

\begin{document}
\let\WriteBookmarks\relax
\def\floatpagepagefraction{1}
\def\textpagefraction{.001}
\shorttitle{Assisted Refinement Network for Camouflaged and Salient Object Detection}
\shortauthors{K. Wang et~al.}

\title [mode = title]{Assisted Refinement Network Based on Channel Information Interaction for Camouflaged and Salient Object Detection}                      
\tnotemark[1]

\tnotetext[1]{This work was supported in part by the Xinjiang Cheonchi Talent Introduction Program, in part by the Xinjiang Efficient Scientific Research Business Funds Project under Grant XJEDU2025J001, and in part by the Xinjiang Graduate Innovation and Practice Program under Grant XJ2025G087.}

\author[1]{Kuan Wang}
\fnmark[1]
\ead{wkuan@stu.xju.edu.cn}

\author[1]{Yanjun Qin}
\fnmark[1]
\ead{qinyanjun@xju.edu.cn}

\author[1]{Mengge Lu}
\ead{107552403966@stu.xju.edu.cn}

\author[1]{Liejun Wang}
\ead{wljxju@xju.edu.cn}

\author[1,2]{Xiaoming Tao}
\cormark[1]
\ead{taoxm@xju.edu.cn}
\ead{taoxm@tsinghua.edu.cn}

\affiliation[1]{organization={School of Computer Science and Technology, Xinjiang University},
                city={Urumqi},
                postcode={830017},
                country={China}}

\affiliation[2]{organization={Department of Electronic Engineering, Tsinghua University},
                city={Beijing},
                postcode={100084},
                country={China}}

\cortext[1]{Corresponding author: Xiaoming Tao.}
\fntext[1]{Equal contribution: Kuan Wang and Yanjun Qin.}


\begin{abstract}
Camouflaged Object Detection (COD) stands as a significant challenge in computer vision, dedicated to identifying and segmenting objects visually highly integrated with their backgrounds. Current mainstream methods have made progress in cross-layer feature fusion, but two critical issues persist during the decoding stage. The first is insufficient cross-channel information interaction within the same-layer features, limiting feature expressiveness. The second is the inability to effectively co-model boundary and region information, making it difficult to accurately reconstruct complete regions and sharp boundaries of objects. To address the first issue, we propose the Channel Information Interaction Module (CIIM), which introduces a horizontal–vertical integration mechanism in the channel dimension. This module performs feature reorganization and interaction across channels to effectively capture complementary cross-channel information. To address the second issue, we construct a collaborative decoding architecture guided by prior knowledge. This architecture generates boundary priors and object localization maps through Boundary Extraction (BE) and Region Extraction (RE) modules, then employs hybrid attention to collaboratively calibrate decoded features, effectively overcoming semantic ambiguity and imprecise boundaries. Additionally, the Multi-scale Enhancement (MSE) module enriches contextual feature representations. Extensive experiments on four COD benchmark datasets validate the effectiveness and state-of-the-art performance of the proposed model. We further transferred our model to the Salient Object Detection (SOD) task and demonstrated its adaptability across downstream tasks, including polyp segmentation, transparent object detection, and industrial and road defect detection. Code and experimental results are publicly available at: \url{https://github.com/akuan1234/ARNet-v2}.


\end{abstract}




\maketitle

\section{Introduction}

Camouflaged Object Detection (COD) is a critical and highly challenging task in computer vision, aiming to accurately identify and localize objects that visually blend with their surroundings~\cite{sinet,stevens2009animal,wang2024depth}. This task extends beyond purely academic research interests, holding significant application value across multiple domains including military reconnaissance~\cite{stevens2009animal}, medical image segmentation (e.g., polyp segmentation)~\cite{fan2020pranet}, industrial defect detection~\cite{fan2023advances}, and agricultural pest and disease monitoring and control~\cite{wang2024depth}. The inherent challenge of COD stems from the evolutionary or design strategies adopted by objects to minimize their distinguishability, making precise segmentation difficult for both biological and artificial visual systems.

\begin{figure*}[!t]
\centering
\includegraphics[width=6.0in]{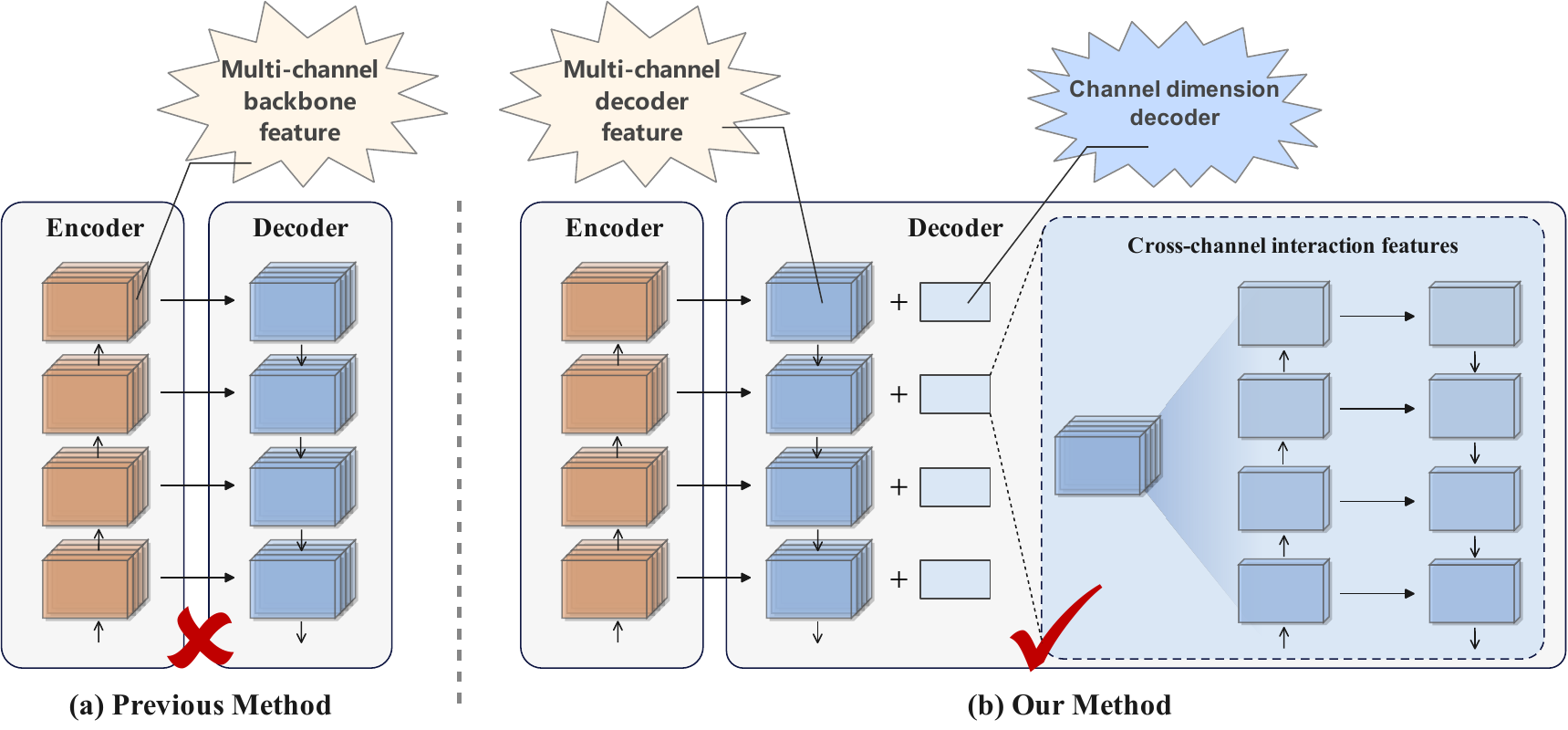}
\caption{Our motivation. (a) Previous methods mainly focus on cross-layer feature fusion, while overlooking information disparities between channels within the same layer. (b) Our method builds upon an Encoder-Decoder architecture designed for cross-layer features, embedding sub-decoders for cross-channel dimensions within each layer's features to fully exploit complementary information across channels.}
\label{fig_1}
\end{figure*}
Current mainstream COD methods primarily focus on cross-layer feature fusion to decode information from different levels of the backbone network~\cite{zoomnet,BGnet,wang2024depth,c2fnet-v2,SARNET}. Despite significant progress, these methods still struggle to achieve precise segmentation in complex scenes. This stems from two limitations in the decoding process: (1) Convolutional neural networks (CNNs) inherently generate multi-channel feature maps, where different channels often represent distinct scene attributes (e.g., texture, boundary, color, or semantic components)~\cite{qiu2021slimconv}. In complex scenarios like camouflaged objects, neglecting information exchange between channels within the same layer can lead to the loss of fine-grained features; (2) The lack of collaborative modeling for boundary and region information results in insufficient boundary accuracy and semantic consistency in prediction outcomes.

To address these limitations, we propose a novel assisted refinement network (ARNet) that fully captures cross-channel complementary information during decoding to generate segmentation results with complete objects and clear boundaries. Specifically, we design a Channel Information Interaction Module (CIIM) employing a dual-dimensional decoding architecture across both cross-layer and cross-channel dimensions. As shown in Fig.~\ref{fig_1}, cross-channel sub-decoders are embedded within each layer of the cross-layer decoder to more deeply mine complementary features across channels. Unlike methods such as SENet~\cite{hu2018squeeze} that merely reweight dominant channels via channel attention, CIIM introduces a structural reorganization paradigm. It decomposes features along the channel dimension and then reorganizes and compresses them along horizontal and vertical axes, thereby explicitly modeling cross-channel dependencies. Additionally, we design Boundary Extraction (BE) and Region Extraction (RE) modules to model prior information from backbone features. The BE module focuses on extracting salient boundary features to generate a boundary prior, while the RE module captures coarse spatial localization to produce an object localization map. To achieve efficient synergy between these modules, we further propose a Hybrid Guided Attention (HGA) module. Within CIIM, HGA dynamically fuses boundary and region guidance through a self-attention mechanism, providing precise boundary and spatial constraints for decoding while maintaining semantic consistency. Finally, a Multi-scale Enhancement (MSE) module is designed to further strengthen the contextual expression capabilities of CIIM's fused features. We conducted comprehensive evaluations on four widely used COD benchmark datasets. Extensive experimental results demonstrate that ARNet significantly outperforms numerous state-of-the-art (SOTA) methods across multiple standard evaluation metrics. To rigorously evaluate the versatility and cross-domain generalization of the proposed method, ARNet was further applied to the closely related task of Salient Object Detection (SOD)~\cite{borji2015salient}, where it exhibited strong adaptability. Furthermore, its effectiveness was validated across diverse downstream applications, including medical image segmentation (polyps), transparent object detection, defect detection, and road crack detection. These experiments clearly demonstrate the model’s robustness, adaptability, and broad applicability across multiple visual domains.

In summary, the main contributions of this paper are threefold:
\begin{itemize}
\item{We propose ARNet, which introduces sub-decoders at the channel dimension to enable dual-dimensional information interaction, explicitly addressing the previously underutilized intra-layer cross-channel information disparity.}
\item{We construct an assistive decoding architecture guided by prior knowledge, with dedicated modules to generate precise boundary priors and object localization maps. We introduce hybrid self-attention to jointly calibrate boundary and spatial localization. Finally, a multi-scale convolutional strategy is employed to enrich the calibrated feature representations.}
\item{We conduct extensive validation on four COD benchmark datasets and successfully transferred our approach to SOD tasks and various downstream applications, demonstrating robust generalization capabilities.}
\end{itemize}

\section{Related Work}

\subsection{Camouflaged Object Detection}

COD aims to identify objects that visually blend into their surroundings, posing an inherent challenge to visual perception systems. With the rapid advancement of deep learning, frameworks based on CNNs~\cite{sinet} and Transformers~\cite{dtinet} have significantly improved overall performance in COD tasks. Current research on supervised learning methods primarily focuses on two core technical approaches: cross-layer feature fusion and assistive information guidance. However, both categories of methods exhibit inherent limitations, necessitating further exploration to achieve more refined and comprehensive segmentation of camouflaged objects.

{\bf{Cross-layer feature fusion strategies:}} Whether based on CNNs or Transformer variants, the feature extraction capabilities of backbone networks exhibit multi-level and multi-scale characteristics. Existing approaches commonly employ multi-layer feature integration strategies, achieving refined segmentation through stepwise decoding from coarse to fine details. Typical approaches include SINet~\cite{sinet}, which proposes a hierarchical framework mimicking the “search-and-identify” mechanism of natural predators; ZoomNet~\cite{zoomnet} and TANet~\cite{TANet}, which further optimize feature representation through multi-scale fusion and progressive enhancement; and SegMar~\cite{segmar} and HitNet~\cite{hitnet}, which refine features through multiple iterations to capture subtle structures and complex textures of objects. Although these approaches have made significant progress in inter-layer information fusion, they generally neglect complementary information exchange between channels within the same layer. Even the few works considering channel relationships (e.g., SENet~\cite{hu2018squeeze}, FPNet~\cite{fpnet}) mostly rely on global weighting methods to recalibrate channel features. This indirect, global modulation approach lacks modeling of explicit and structured dependencies between channels, limiting the full realization of feature expression capabilities.

{\bf{Assistive information guidance strategies:}} To mitigate the extremely low-contrast between camouflaged objects and their backgrounds, some studies introduce explicit prior information to guide the network in learning more discriminative features. Among these, boundary information is considered the most representative guiding signal. Typical works such as BGNet~\cite{BGnet} and its improved version~\cite{BgNetchen2022boundary} enhance structural consistency by introducing boundary branches or dual-decoder structures; Additionally, methods have leveraged assisted constraints such as category priors~\cite{zhang2024cgcod}, gradient features~\cite{ji2023deep}, or certainty maps~\cite{lai2025certainty} to enhance semantic discrimination. However, these approaches generally decouple boundary and region modeling, employing independent branches or loose fusion, making it challenging to achieve synergistic optimization. This non-cooperative guidance mechanism often leads to segmentation artifacts such as blurred boundaries and incomplete regions. Moreover, assisting information is frequently not deeply embedded into the main decoding process, preventing its potential complementary constraints from being fully leveraged. Achieving cooperative decoding of boundary structure and region localization remains a critical challenge for improving the integrity and accuracy of camouflaged objects.

\subsection{Salient Object Detection}

SOD aims to identify objects or regions in images that most attract human visual attention. Although its task definition appears opposite to COD, the two share significant commonalities in underlying technologies and core challenges~\cite{zhu2021inferring}. The technical evolution path of modern deep learning-driven SOD methods exhibits remarkable convergence with COD.

To achieve precise segmentation, mainstream SOD methods commonly employ guidance strategies centered on boundary and structure information, such as incorporating auxiliary boundary detection branches, designing boundary-oriented loss functions, or implementing boundary-aware modules~\cite{feng2019attentive,wu2019stacked}. Concurrently, researchers focus on designing sophisticated multi-scale feature fusion and attention mechanisms~\cite{wu2019cascaded} to balance high-level semantic information with low-level spatial details. In recent years, Transformer architectures~\cite{xie2021segmenting} have been introduced to enhance global context modeling. These technical strategies, shared with COD, reveal that both tasks fundamentally challenge the precise separation of foreground and background, heavily relying on a model's ability to discern fine-grained features.

Therefore, the core limitations we revealed in the COD task are equally significant in the SOD task. A network designed to tackle the more challenging COD task, capable of fully leveraging complementary information across channels and enabling assisted guided decoding, should demonstrate superior performance in the SOD task. Based on this, we transferred ARNet to the SOD task to validate the universality of its design and the model's generalization capability. This process also provides robust theoretical support for its cross-task adaptability.

\subsection{Downstream Applications Related to COD}

The core capability of COD lies in distinguishing objects with subtle feature differences from complex or homogeneous backgrounds, endowing it with broad practical applications beyond academic research. In recent years, with the rapid advancement of COD technology, it has been successfully applied to various visual tasks, including medical image segmentation (e.g., polyp segmentation~\cite{ji2021progressively}, COVID-19 lung lesion segmentation~\cite{fan2020inf}), industrial surface defect detection, and transparent object detection~\cite{fan2023advances}. A common characteristic of these tasks is that target objects often exhibit blurred boundaries and low-contrast with their surroundings, aligning closely with the inherent challenges of COD tasks.

Among numerous downstream applications, polyp segmentation has garnered significant attention due to its critical clinical value and close association with colorectal cancer (CRC)~\cite{ji2021progressively}. In colonoscopy images, the color and texture of polyps often closely resemble those of normal intestinal mucosa, making them a classic “camouflaged” object in medical imaging. Accurate polyp segmentation is crucial for the early diagnosis and prevention of colorectal cancer. Similar to the development trajectory of COD/SOD, polyp segmentation techniques have evolved from reliance on manual features~\cite{manual} to the adoption of deep learning approaches. Modern approaches, such as various models based on fully convolutional networks (FCNs) (e.g., U-Net++~\cite{zhou2019unet++}, PraNet~\cite{fan2020pranet}), also emphasize the utilization of boundary information~\cite{sfa} and advanced attention mechanisms~\cite{ji2021progressively} to enhance segmentation accuracy.

This striking similarity in challenges and solutions demonstrates that advanced models designed for COD problems possess significant potential for transfer learning and tackling such downstream tasks. Therefore, this paper adopts polyp segmentation and similar tasks as critical validation platforms to assess the effectiveness, generalization capability, and versatility of our proposed ARNet in real-world, challenging applications.

\begin{figure*}[!t]
\centering
\includegraphics[width=\textwidth]{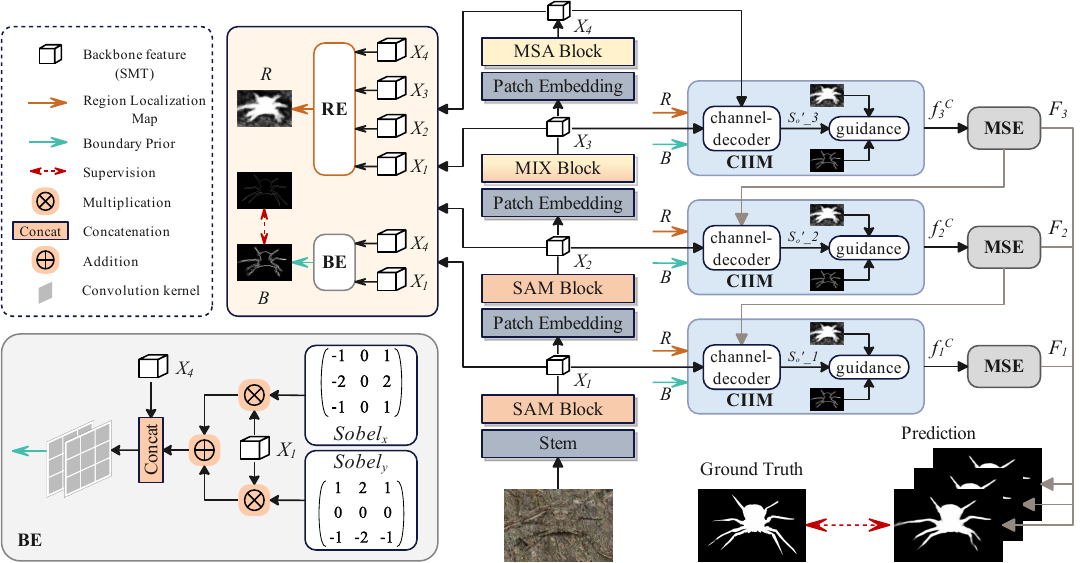}
\caption{The overall architecture of the proposed ARNet, primarily comprising four core modules: Channel Information Interaction Module (CIIM), Boundary Extraction (BE) module, Region Extraction (RE) module, and Multi-scale Enhancement (MSE) module. Input images are extracted through a backbone network to obtain multi-level features, which undergo dual-dimensional decoding in the CIIM module. Subsequently, the boundary prior and region localization map generated by RE and BE are utilized for assisted guidance. Finally, the MSE employs a multi-scale convolutional strategy to expand the receptive field, enhance feature expression capabilities, and guide low-level feature decoding. The feature $F_{1}$ output from the final MSE layer serves as the final prediction result.}
\label{fig_2}
\end{figure*}


\section{Methodology}

\subsection{Overview}
The overall architecture of ARNet is illustrated in Fig.~\ref{fig_2}, which mainly consists of three key components: a feature extraction backbone, a main decoding network, and auxiliary sub-tasks.
The feature extraction backbone adopts the Scale-aware Modulation Transformer (SMT)~\cite{lin2023scale}, which surpasses PVTv2~\cite{wang2022pvt} in terms of its scale-aware modulation mechanism, efficient fusion of ConvNet and Transformer advantages, and overall performance.
Building upon the extracted multi-scale features $X_i$ ($i \in {1, 2, 3, 4}$), the CIIM and MSE modules form the main decoding network, which performs feature decoding across both cross-channel and cross-layer dimensions.
Finally, the RE and BE modules serve as auxiliary sub-tasks, generating semantic priors for region and boundary information from the backbone features, thereby facilitating the decoding process within CIIM.
The entire network adopts a top-down refinement strategy, where high-level features progressively guide lower-level ones to decode and refine the multi-level representations extracted from the backbone.
In addition, multi-level supervision is applied to both the boundary priors and the MSE-generated feature maps at each layer, while the refined features from the final stage are used to produce the final prediction results.

\subsection{Region Extraction Module}

To maintain semantic consistency during the layer-by-layer feature decoding, our designed RE module provides assisted guidance for the primary regions of the features. Furthermore, to prevent the loss of local information in the low-level backbone features, we aggregate all four levels of features in a top-down manner based on their semantic hierarchy. We prioritize the fusion of backbone features with smaller differences between adjacent layers, ultimately generating a unified object localization map $R$ for coarse spatial localization of camouflaged objects. The specific workflow is as follows:
\begin{equation}\begin{cases}X_{34} =C_{3\times3}BR(\mathrm{Cat}(u_\uparrow^2(C_{1 \times 1}BR(X_4)),C_{1\times1}BR(X_3))) \\X_{234}=C_{3\times3}BR(\mathrm{Cat}(u_\uparrow^2(X_{34}),C_{1\times1}BR(X_2)))\\R=C_{3\times3}BR(\mathrm{Cat}(u_\uparrow^2(X_{234}),C_{1\times1}BR(X_1)))\end{cases}\end{equation}
where $C_{i \times i}BR(\cdot)$ represents convolution units consisting of $i \times i$ convolution, batch normalization, and $ReLU(\cdot)$ activation functions, $u_\uparrow^2(\cdot)$ represents upsampling factor of 2, and $Cat(\cdot)$ represents concatenation operations. To ensure the same dimensionality, we sequentially performed upsampling and channel adjustment on the high-level features.

\subsection{Boundary Extraction Module}

In COD, objects typically exhibit low-contrast boundaries against the background, making boundary cues crucial for feature decoding. Therefore, in addition to region localization guidance, we introduce a boundary prior guidance. Specifically, we generate the boundary prior by leveraging the distinct boundary information in the high-resolution feature $X_1$ and the differences between it and the low-resolution feature $X_4$. To enhance the boundary information in $X_1$ and amplify the differences with $X_4$, we compute gradient maps of $X_1$ using Sobel operators in both horizontal and vertical directions~\cite{kanopoulos1988design}, then fuse them to obtain the boundary-enhanced $X_{1}^{\prime}$. The Sobel convolution factors and calculation formula are as follows:
\begin{equation}Sobel_x=
\begin{bmatrix}
-1 & 0 & +1 \\
-2 & 0 & +2 \\
-1 & 0 & +1
\end{bmatrix}, Sobel_y=
\begin{bmatrix}
+1 & +2 & +1 \\
0 & 0 & 0 \\
-1 & -2 & -1
\end{bmatrix}\end{equation}
\begin{equation}X_{1}^{\prime}=\sqrt{\left(Sobel_{x}\times X_{1}\right)^{2}+\left(Sobel_{y}\times X_{1}\right)^{2}}\end{equation}
Then upsample $X_4$ to match the resolution of $X_{1}^{\prime}$, concatenate the two feature maps, fuse them through convolution blocks to learn boundary variations, and obtain the precise boundary prediction map $B$ through supervision by the boundary ground truth map. The specific workflow can be summarized as follows:
\begin{equation}B=Sig((C_{3\times3}BR)_2(Cat(X_1^{\prime},u_\uparrow^8(X_4))))\end{equation}
where $Sig(\cdot)$ denotes the Sigmoid activation function, and the subscript 2 indicates two sequential convolutional blocks.

\begin{figure*}[!t]
\centering
\includegraphics[width=6.5in]{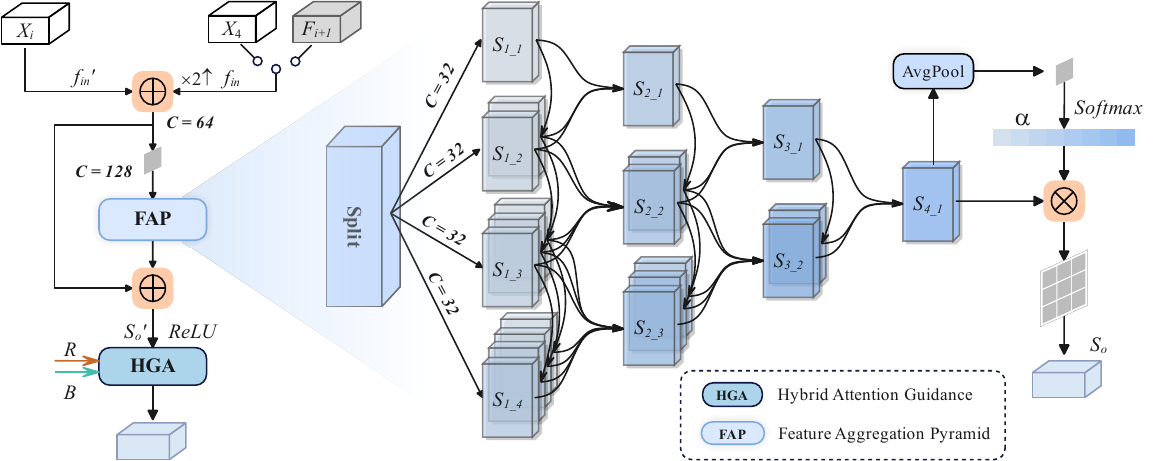}
\caption{Internal architecture of CIIM, embedding two core components: FAP and HGA. The former decodes features at the channel dimension, while the latter assists feature decoding through hybrid self-attention.}
\label{fig_3}
\end{figure*}

\subsection{Channel Information Interaction Module}

Fig.~\ref{fig_3} shows the internal details of CIIM. Within CIIM, the input small-scale feature $f_{in}$ undergoes bilinear interpolation to double its scale before being added to the adjacent feature $f_{in}^{\prime}$. To preserve sufficient channel information, a convolution block with a kernel size of 1 integrates the features, expanding the number of channels from 64 to 128. Subsequently, the Feature Aggregation Pyramid (FAP) decodes the feature along the channel dimension. To balance channel information interaction against model complexity, the feature map is partitioned uniformly into four segments along the channel dimension, yielding $S_{1_{-}i}, i \in \{1,2,3,4\}$. Each feature now has 32 channels. This process can be defined as follows:
\begin{equation}S=u_{\uparrow}^2\left(f_{in}\right)\oplus f_{in}^{\prime}\end{equation}
\begin{equation}S_{1_{-}1},...,S_{1_{-}4}=Split(C_{1\times1}BR(S))\end{equation}
where $Split(\cdot)$ denotes a channel splitting operation, $\oplus$ is an element-wise summation.
\begin{figure*}[!t]
\centering
\includegraphics[width=6.5in]{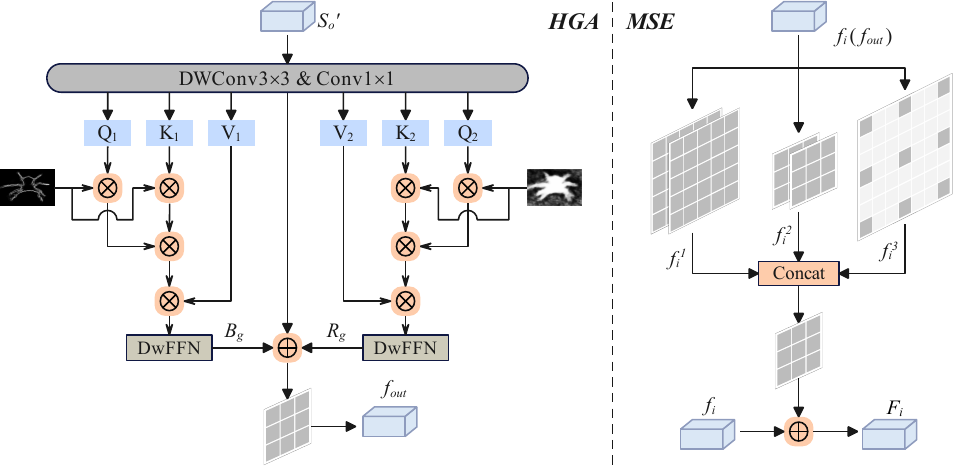}
\caption{The left and right sides show the internal architecture of HGA and MSE, respectively.}
\label{fig_4}
\end{figure*}

In FAP, the segmented feature maps are fused and contracted layer by layer from top to bottom and left to right, fully extracting hidden information across each channel. Specifically, for $S_{1_{-}i}, i \in \{1,2,3,4\}$, vertical layer-by-layer fusion is first performed in a cascading manner. We combine $S_{1_{-}1}$ and $S_{1_{-}2}$ through a channel concatenation and convolution operation to obtain the interacted features. This result is then interacted with $S_{1_{-}3}$ in the same manner to obtain aggregated features, which are subsequently interacted with $S_{1_{-}4}$. This constitutes the vertical fusion process at the first layer. Horizontal fusion of adjacent cross-channel features forms the second layer. The implementation process is as follows: the interaction result of $S_{1_{-}1}$ and $S_{1_{-}2}$ is convolved with $S_{1_{-}1}$ using the same concatenation-convolution operation to yield $S_{2_{-}1}$. This process iterates sequentially through four layers, aggregating segmented features into $S_{4_{-}1}$. The interaction at each layer can be abstracted into the following recursive formula:
\begin{equation}
    F(x_1,x_2)=C_{3\times3}BR(Cat(x_1,x_2))
\end{equation}
\begin{equation}
\begin{cases}
S_{m,k} = F( C_{(m-1),k}, F( C_{(m-1),k}, S_{(m-1),(k + 1)} ) ) \\
C_{(m-1),k} = \begin{cases}
S_{(m-1),1} &  k = 1 \\
F( C_{(m-1),(k - 1)}, S_{(m-1),k}) &  k = 2,3,4
\end{cases}
\end{cases}
\end{equation}
Among these, the $F(x_1,x_2)$ function represents a combination of cascade and convolution operations on two input features. Separately, the symbols $S_{m,k}$ and $C_{m-1,k}$ denote $S_{m_{-}k}$ (for $m \geq 2$) and an intermediate variable, respectively.

We introduce a gating mechanism to weight each channel of the feature map, enhancing the adaptive selection capability of different channel features. First, the gating signal $\alpha$ is generated through global average pooling, convolution, and Softmax operations. Next, the gating signal is multiplied element-wise with the input feature map to achieve channel-level weighting. The weighted feature map is then fused through a convolution layer, ultimately producing the output feature $S_{o}$. The specific workflow can be represented as follows:
\begin{equation}S_{o}=C_{3\times3}BR(S_{4_{-}1}\otimes Softmax(C_{1\times1}(Avg(S_{4_{-}1}))))\end{equation}
where $\otimes$ denotes element-wise multiplication and $Avg(\cdot)$ denotes global average pooling.

Residual learning on $S_{o}$ yields the enhanced features $S_{o}^{\prime}$ after channel-wise decoding. Finally, we designed the HGA module to provide dual guidance to $S_{o}^{\prime}$ from both boundary and region perspectives. The internal schematic of HGA is shown in Fig.~\ref{fig_4}. We introduce and refine a self-attention mechanism. First, the boundary prior and object localization map are embedded into a dual-path attention mechanism, focusing on spatial and global feature guidance. Subsequently, local features are preserved via a Depth-wise Separable Convolution (DwFFN)~\cite{dwffn}, while enhanced features are fused through residual connections with $S_{o}^{\prime}$. Specifically, $S_{o}^{\prime}$ generates \textbf{Q}, \textbf{K}, and \textbf{V} through $1\times1$ convolutions. In each improved attention mechanism, the guidance map injects features into \textbf{Q} and \textbf{K} via element-wise multiplication, then fuses them following the attention mechanism. The resulting features are fed into DwFFN to obtain the guidance features $B_{g}$ and $R_{g}$ for each channel. Finally, these are fused with $S_{o}^{\prime}$ to produce the output feature $f_{out}$, as illustrated in the following flow:
\begin{equation}
\begin{cases}
\mathbf{Q}_{1},\mathbf{K}_{1},\mathbf{V}_{1},\mathbf{Q}_{2},\mathbf{K}_{2},\mathbf{V}_{2}=DWC_{3\times3}(C_{1\times1}(S_{o}^{\prime})) \\ B_{g}=DwFFN((Q_{1}\otimes B)\otimes(K_{1}\otimes B)\otimes V_{1})\\R_{g}=DwFFN((Q_{2}\otimes R)\otimes(K_{2}\otimes R)\otimes V_{2})\\f_{out}=C_{3\times3}BR(B_{g}\oplus R_{g}\oplus S_{o}^{\prime})
\end{cases}
\end{equation}

\begin{table*}[!t]
\caption{Quantitative evaluation against SOTA methods on COD10K, CAMO, NC4K, and CHAMELEON datasets. The best performing results are marked in \textcolor{red}{red}, while the second best results are highlighted in \textcolor{blue}{blue}.}
\setlength{\tabcolsep}{3pt}
\renewcommand{\arraystretch}{1} 
\label{tab1}
\centering
{\fontsize{6}{8}\selectfont 
\resizebox{\linewidth}{!}{
\begin{tabular}{c|c|c|c|c|c}
\hline
\multirow{2}{*}{Method} &\multirow{2}{*}{Source} & COD10K-Test \cite{sinet} & CAMO-Test \cite{camo} & NC4K \cite{nc4k} & CHAMELEON \cite{chameleon}\\ \cline{3-6}
 & & \begin{tabular}{cccc}$S_{\alpha}$$\uparrow$& $E_{\phi}$$\uparrow$ &$F_{\beta}^{\omega}$$\uparrow$ &$\mathcal{M}$$\downarrow$ \end{tabular}& \begin{tabular}{cccc}$S_{\alpha}$$\uparrow$& $E_{\phi}$$\uparrow$ &$F_{\beta}^{\omega}$$\uparrow$ &$\mathcal{M}$$\downarrow$ \end{tabular} &\begin{tabular}{cccc}$S_{\alpha}$$\uparrow$& $E_{\phi}$$\uparrow$ &$F_{\beta}^{\omega}$$\uparrow$ &$\mathcal{M}$$\downarrow$ \end{tabular} & \begin{tabular}{cccc}$S_{\alpha}$$\uparrow$& $E_{\phi}$$\uparrow$ &$F_{\beta}^{\omega}$$\uparrow$ &$\mathcal{M}$$\downarrow$ \end{tabular}
 \\ \hline

SINet-v2 \cite{sinet-v2} & TPAMI22 & \begin{tabular}{cccc}0.815 & 0.887 & 0.680 & 0.037\end{tabular} & \begin{tabular}{cccc}0.820 & 0.882 & 0.743 & 0.070\end{tabular} & \begin{tabular}{cccc}\raisebox{5\height}{\rule{2.26em}{0.4pt}} & \raisebox{5\height}{\rule{2.26em}{0.4pt}} & \raisebox{5\height}{\rule{2.26em}{0.4pt}} & \raisebox{5\height}{\rule{2.26em}{0.4pt}}\end{tabular} & \begin{tabular}{cccc}0.888 & 0.942 & 0.816 & 0.030\end{tabular} \\
C2FNet-v2 \cite{c2fnet-v2} & TCSVT22 & \begin{tabular}{cccc}0.811 & 0.891 & 0.691 & 0.036\end{tabular} & \begin{tabular}{cccc}0.800 & 0.869 & 0.730 & 0.077\end{tabular} & \begin{tabular}{cccc}\raisebox{5\height}{\rule{2.26em}{0.4pt}} & \raisebox{5\height}{\rule{2.26em}{0.4pt}} & \raisebox{5\height}{\rule{2.26em}{0.4pt}} & \raisebox{5\height}{\rule{2.26em}{0.4pt}}\end{tabular} & \begin{tabular}{cccc}0.893 & 0.947 & 0.845 & 0.028\end{tabular} \\
ZoomNet \cite{zoomnet} & CVPR22 & \begin{tabular}{cccc}0.838 & 0.911 & 0.729 & 0.029\end{tabular} & \begin{tabular}{cccc}0.820 & 0.892 & 0.752 & 0.066\end{tabular} & \begin{tabular}{cccc}0.853 & 0.912 & 0.784 & 0.043\end{tabular} & \begin{tabular}{cccc}0.902 & \raisebox{5\height}{\rule{2.26em}{0.4pt}} & 0.845 & 0.023\end{tabular} \\
BGNet \cite{BGnet} & IJCAI22 & \begin{tabular}{cccc}0.831 & 0.901 & 0.722 & 0.033\end{tabular} & \begin{tabular}{cccc}0.812 & 0.870 & 0.749 & 0.073\end{tabular} & \begin{tabular}{cccc}0.851 & 0.907 & 0.788 & 0.044\end{tabular} & \begin{tabular}{cccc}\raisebox{5\height}{\rule{2.26em}{0.4pt}} & \raisebox{5\height}{\rule{2.26em}{0.4pt}} & \raisebox{5\height}{\rule{2.26em}{0.4pt}} & \raisebox{5\height}{\rule{2.26em}{0.4pt}}\end{tabular} \\
FEDER \cite{feder} & CVPR23 & \begin{tabular}{cccc}0.844 & 0.911 & 0.748 & 0.029\end{tabular} & \begin{tabular}{cccc}0.836 & 0.897 & 0.807 & 0.066\end{tabular} & \begin{tabular}{cccc}0.862 & 0.913 & 0.824 & 0.042\end{tabular} & \begin{tabular}{cccc}0.906 & 0.959 & 0.869 & 0.023\end{tabular}\\
FSPNet \cite{fspnet} & CVPR23 & \begin{tabular}{cccc}0.851 & 0.895 & 0.735 & 0.026\end{tabular} & \begin{tabular}{cccc}0.856 & 0.899 & 0.799 & 0.050\end{tabular} & \begin{tabular}{cccc}0.879 & 0.915 & 0.816 & 0.035\end{tabular} & \begin{tabular}{cccc}\raisebox{5\height}{\rule{2.26em}{0.4pt}} & \raisebox{5\height}{\rule{2.26em}{0.4pt}} & \raisebox{5\height}{\rule{2.26em}{0.4pt}} & \raisebox{5\height}{\rule{2.26em}{0.4pt}}\end{tabular} \\
OAFormer \cite{oaformer} & ICME23 & \begin{tabular}{cccc}0.860 & 0.927 & 0.773 & 0.025\end{tabular} & \begin{tabular}{cccc}0.866 & 0.924 & 0.826 & 0.048\end{tabular} & \begin{tabular}{cccc}0.883 & 0.934 & 0.837 & 0.033\end{tabular} & \begin{tabular}{cccc}0.904 & \textcolor{blue}{0.960} & 0.858 & 0.023\end{tabular} \\
FPNet \cite{fpnet} & MM23 & \begin{tabular}{cccc}0.850 & 0.913 & 0.748 & 0.029\end{tabular} & \begin{tabular}{cccc}0.852 & 0.905 & 0.806 & 0.056\end{tabular} & \begin{tabular}{cccc}\raisebox{5\height}{\rule{2.26em}{0.4pt}} & \raisebox{5\height}{\rule{2.26em}{0.4pt}} & \raisebox{5\height}{\rule{2.26em}{0.4pt}} & \raisebox{5\height}{\rule{2.26em}{0.4pt}}\end{tabular} & \begin{tabular}{cccc}0.914 & \raisebox{5\height}{\rule{2.26em}{0.4pt}} & 0.856 & 0.022\end{tabular} \\
SARNet \cite{SARNET} & TCSVT23 & \begin{tabular}{cccc}0.864 & 0.931 & 0.777 & 0.024\end{tabular} & \begin{tabular}{cccc}0.868 & 0.927 & 0.828 & 0.047\end{tabular} & \begin{tabular}{cccc}0.886 & 0.937 & 0.842 & 0.032\end{tabular} & \begin{tabular}{cccc}0.912 & 0.957 & 0.871 & 0.021\end{tabular} \\
DINet \cite{dinet} & TMM24 & \begin{tabular}{cccc}0.832 & 0.903 & \raisebox{5\height}{\rule{2.26em}{0.4pt}} & 0.031\end{tabular} & \begin{tabular}{cccc}0.821 & 0.874 & \raisebox{5\height}{\rule{2.26em}{0.4pt}} & 0.068\end{tabular} & \begin{tabular}{cccc}0.856 & 0.909 & \raisebox{5\height}{\rule{2.26em}{0.4pt}} & 0.043\end{tabular} & \begin{tabular}{cccc}\raisebox{5\height}{\rule{2.26em}{0.4pt}} & \raisebox{5\height}{\rule{2.26em}{0.4pt}} & \raisebox{5\height}{\rule{2.26em}{0.4pt}} & \raisebox{5\height}{\rule{2.26em}{0.4pt}}\end{tabular} \\
CamoFormer \cite{camoformer} & TPAMI24 & \begin{tabular}{cccc}0.872 & 0.934 & 0.793 & \textcolor{blue}{0.022}\end{tabular} & \begin{tabular}{cccc}\textcolor{blue}{0.878} & \textcolor{blue}{0.934} & 0.839 & \textcolor{blue}{0.044}\end{tabular} & \begin{tabular}{cccc}\textcolor{blue}{0.893} & \textcolor{blue}{0.940} & 0.850 & \textcolor{blue}{0.030}\end{tabular} & \begin{tabular}{cccc}\raisebox{5\height}{\rule{2.26em}{0.4pt}} & \raisebox{5\height}{\rule{2.26em}{0.4pt}} & \raisebox{5\height}{\rule{2.26em}{0.4pt}} & \raisebox{5\height}{\rule{2.26em}{0.4pt}}\end{tabular} \\
PRNet \cite{prnet} & TCSVT24 & \begin{tabular}{cccc}0.874 & 0.937 & \raisebox{5\height}{\rule{2.26em}{0.4pt}} & \textcolor{blue}{0.022}\end{tabular} & \begin{tabular}{cccc}0.872 & 0.922 & \raisebox{5\height}{\rule{2.26em}{0.4pt}} & 0.050\end{tabular} & \begin{tabular}{cccc}0.891 & 0.933 & \raisebox{5\height}{\rule{2.26em}{0.4pt}} & 0.031\end{tabular} & \begin{tabular}{cccc}0.914 & \textcolor{blue}{0.960} & \raisebox{5\height}{\rule{2.26em}{0.4pt}} & \textcolor{blue}{0.020}\end{tabular} \\
DRFNet \cite{drfnet} & TCSVT24 & \begin{tabular}{cccc}0.869 & 0.936 & 0.792 & 0.023\end{tabular} & \begin{tabular}{cccc}0.868 & 0.925 & 0.832 & 0.047\end{tabular} & \begin{tabular}{cccc}0.887 & 0.939 & 0.846 & 0.031\end{tabular} & \begin{tabular}{cccc}\textcolor{blue}{0.918} & 0.959 & \textcolor{blue}{0.880} & \textcolor{red}{0.019}\end{tabular} \\
RISNet \cite{wang2024depth} & CVPR24& \begin{tabular}{cccc}0.873 & 0.931 & 0.799 & 0.025\end{tabular}
 & \begin{tabular}{cccc}0.870 & 0.922 & 0.827 & 0.050\end{tabular}
& \begin{tabular}{cccc}0.882 & 0.925 & 0.834 & 0.037\end{tabular} & \begin{tabular}{cccc}\raisebox{5\height}{\rule{2.26em}{0.4pt}} & \raisebox{5\height}{\rule{2.26em}{0.4pt}} & \raisebox{5\height}{\rule{2.26em}{0.4pt}} & \raisebox{5\height}{\rule{2.26em}{0.4pt}}\end{tabular} \\
CSFIN \cite{LI2025126451} & ESWA25 & \begin{tabular}{cccc}0.868 & 0.930 & 0.780 & 0.023\end{tabular} & \begin{tabular}{cccc}0.876 & 0.929 & 0.832 & 0.047\end{tabular} & \begin{tabular}{cccc}0.890 & 0.937 & 0.842 & 0.031\end{tabular} & \begin{tabular}{cccc}\raisebox{5\height}{\rule{2.26em}{0.4pt}} & \raisebox{5\height}{\rule{2.26em}{0.4pt}} & \raisebox{5\height}{\rule{2.26em}{0.4pt}} & \raisebox{5\height}{\rule{2.26em}{0.4pt}}\end{tabular} \\
PRBE-Net \cite{prbenet} & TMM25 & \begin{tabular}{cccc}0.867 & 0.932 & 0.793 & \textcolor{red}{0.021}\end{tabular} & \begin{tabular}{cccc}0.876 & 0.928 & 0.837 & 0.045\end{tabular} & \begin{tabular}{cccc}0.887 & 0.931 & 0.845 & 0.031\end{tabular} & \begin{tabular}{cccc}\textcolor{blue}{0.918} & 0.951 & 0.878 & \textcolor{blue}{0.020}\end{tabular} \\
RUN \cite{run} & ICML25 & \begin{tabular}{cccc}\textcolor{blue}{0.878} & \textcolor{red}{0.941} & \textcolor{red}{0.810} & \textcolor{red}{0.021}\end{tabular} & \begin{tabular}{cccc}0.877 & \textcolor{blue}{0.934} & \textcolor{red}{0.861} & 0.045\end{tabular} & \begin{tabular}{cccc}0.892 & \textcolor{blue}{0.940} & \textcolor{red}{0.868} & \textcolor{blue}{0.030}\end{tabular} & \begin{tabular}{cccc}0.916 & 0.958 & 0.877 & 0.021\end{tabular} \\
\hline
ARNet & 2025 & \begin{tabular}{cccc}\textcolor{red}{0.883} & \textcolor{blue}{0.938} & \textcolor{blue}{0.808} & \textcolor{red}{0.021}\end{tabular} & \begin{tabular}{cccc}\textcolor{red}{0.890} & \textcolor{red}{0.937} & \textcolor{blue}{0.853} & \textcolor{red}{0.042}\end{tabular} & \begin{tabular}{cccc}\textcolor{red}{0.899} & \textcolor{red}{0.942} & \textcolor{blue}{0.857} & \textcolor{red}{0.029}\end{tabular} & \begin{tabular}{cccc}\textcolor{red}{0.921} & \textcolor{red}{0.964} & \textcolor{red}{0.883} & \textcolor{red}{0.019}\end{tabular} \\ \hline
\end{tabular}
}
}
\end{table*}

\begin{table*}[!t]
\caption{Quantitative evaluation of SOTA methods on five SOD datasets using ARNet. The best performing results are marked in \textcolor{red}{red}, while the second best results are highlighted in \textcolor{blue}{blue}.}
\setlength{\tabcolsep}{3pt}
\renewcommand{\arraystretch}{1} 
\label{tab2}
\centering
{\fontsize{6}{8}\selectfont 
\resizebox{\linewidth}{!}{
\begin{tabular}{c|c|c|c|c|c|c}
\hline
\multirow{2}{*}{Method}  &\multirow{2}{*}{Source}  & PASCAL-S\cite{pascal}                        & ECSSD\cite{ecssd}                            & HKU-IS\cite{hku-is}   &   DUT-OMRON~\cite{dut-om} & DUTS-TE~\cite{duts}                   \\ \cline{3-7} 
    & &  \begin{tabular}{cccc}$S_{\alpha}$$\uparrow$& $E_{\phi}$$\uparrow$ &$F_{\beta}^{\omega}$$\uparrow$ &$\mathcal{M}$$\downarrow$ \end{tabular}    &               \begin{tabular}{cccc}$S_{\alpha}$$\uparrow$& $E_{\phi}$$\uparrow$ &$F_{\beta}^{\omega}$$\uparrow$ &$\mathcal{M}$$\downarrow$ \end{tabular} &               \begin{tabular}{cccc}$S_{\alpha}$$\uparrow$& $E_{\phi}$$\uparrow$ &$F_{\beta}^{\omega}$$\uparrow$ &$\mathcal{M}$$\downarrow$ \end{tabular} &               \begin{tabular}{cccc}$S_{\alpha}$$\uparrow$& $E_{\phi}$$\uparrow$ &$F_{\beta}^{\omega}$$\uparrow$ &$\mathcal{M}$$\downarrow$ \end{tabular} &\begin{tabular}{cccc}$S_{\alpha}$$\uparrow$& $E_{\phi}$$\uparrow$ &$F_{\beta}^{\omega}$$\uparrow$ &$\mathcal{M}$$\downarrow$ \end{tabular} 
    \\ \hline

DNA \cite{dna} & TCybern21 & \begin{tabular}{cccc}0.837 & 0880 & 0.798 & 0.074\end{tabular} & \begin{tabular}{cccc}0.852 & 0.905 & 0.806 & 0.056\end{tabular} & \begin{tabular}{cccc}0.913 & 0.955 & 0.898& 0.028\end{tabular} 
& \begin{tabular}{cccc}0.826 & 0.863 & 0.735 & 0.056\end{tabular}  
    & \begin{tabular}{cccc}0.872 & 0.908 & 0.814 & 0.040\end{tabular}\\
VST \cite{vst} & ICCV21     & \begin{tabular}{cccc}0.872 & 0.898 & \raisebox{5\height}{\rule{2.26em}{0.4pt}} & 0.062\end{tabular} & \begin{tabular}{cccc}0.932 & 0.952 & \raisebox{5\height}{\rule{2.26em}{0.4pt}} & 0.034\end{tabular} & \begin{tabular}{cccc}0.928 & 0.953 & \raisebox{5\height}{\rule{2.26em}{0.4pt}} & 0.030\end{tabular} 
& \begin{tabular}{cccc}0.850 & 0.872 & \raisebox{5\height}{\rule{2.26em}{0.4pt}} & 0.058\end{tabular}  
    & \begin{tabular}{cccc}0.896 & 0.920 & \raisebox{5\height}{\rule{2.26em}{0.4pt}} & 0.037\end{tabular}\\
CFNet \cite{cfnet} & PR22 & \begin{tabular}{cccc}\textcolor{blue}{0.877} & \textcolor{blue}{0.916} & \raisebox{5\height}{\rule{2.26em}{0.4pt}} & 0.055\end{tabular} & \begin{tabular}{cccc}0.937 & 0.960 & \raisebox{5\height}{\rule{2.26em}{0.4pt}} & 0.027\end{tabular} & \begin{tabular}{cccc}\textcolor{blue}{0.930} & 0.962 & \raisebox{5\height}{\rule{2.26em}{0.4pt}} & 0.025\end{tabular} 
& \begin{tabular}{cccc}0.854 & 0.884 & \raisebox{5\height}{\rule{2.26em}{0.4pt}} & 0.052\end{tabular}  
    & \begin{tabular}{cccc}0.909 & 0.941 & \raisebox{5\height}{\rule{2.26em}{0.4pt}} & 0.030\end{tabular}\\
DPNet \cite{dpnet} & TIP22  & \begin{tabular}{cccc}\textcolor{blue}{0.877} & 0.907 & \raisebox{5\height}{\rule{2.26em}{0.4pt}} & \textcolor{blue}{0.054}\end{tabular} & \begin{tabular}{cccc}0.931 & 0.952     & \raisebox{5\height}{\rule{2.26em}{0.4pt}} & 0.031\end{tabular} & \begin{tabular}{cccc}\textcolor{red}{0.934} & \textcolor{blue}{0.965} & \raisebox{5\height}{\rule{2.26em}{0.4pt}} & \textcolor{red}{0.023}\end{tabular} 
& \begin{tabular}{cccc}0.854 & 0.878 & \raisebox{5\height}{\rule{2.26em}{0.4pt}} & 0.049\end{tabular}  
    & \begin{tabular}{cccc}0.912 & \textcolor{blue}{0.943} & \raisebox{5\height}{\rule{2.26em}{0.4pt}} & 0.028\end{tabular}\\
ICON \cite{icon} & TPAMI23  & \begin{tabular}{cccc}0.862 & 0.894 & 0.823 & 0.065\end{tabular} & \begin{tabular}{cccc}0.929 & 0.954 & 0.918 & 0.032\end{tabular} & \begin{tabular}{cccc}0.920 & 0.959 & 0.902 & 0.029\end{tabular} 
& \begin{tabular}{cccc}0.844 & 0.879 & 0.761 & 0.057\end{tabular}  
    & \begin{tabular}{cccc}0.889 & 0.919 & 0.837 & 0.037\end{tabular}\\
MENet \cite{menet} & CVPR23  & \begin{tabular}{cccc}0.872 & 0.905 & \raisebox{5\height}{\rule{2.26em}{0.4pt}} & \textcolor{blue}{0.054}\end{tabular} & \begin{tabular}{cccc}0.928 & 0.952 & \raisebox{5\height}{\rule{2.26em}{0.4pt}} & 0.031\end{tabular} & \begin{tabular}{cccc}0.927 & 0.960 & \raisebox{5\height}{\rule{2.26em}{0.4pt}} & 0.023\end{tabular} 
& \begin{tabular}{cccc}0.850 & 0.871 & \raisebox{5\height}{\rule{2.26em}{0.4pt}} & \textcolor{blue}{0.045}\end{tabular}  
    & \begin{tabular}{cccc}0.905 & 0.938 & \raisebox{5\height}{\rule{2.26em}{0.4pt}} & 0.028\end{tabular}\\
MPTNet \cite{mptnet}        &        ESWA24   & \begin{tabular}{cccc}0.862 & 0.901 & 0.817 & 0.063\end{tabular}  
    & \begin{tabular}{cccc}0.930 & 0.956 & \textcolor{blue}{0.920} & 0.032\end{tabular}  
    & \begin{tabular}{cccc}0.922 & 0.954 & \textcolor{blue}{0.903} & 0.029\end{tabular} 
    & \begin{tabular}{cccc}0.846 & 0.878 & 0.762 & 0.056\end{tabular}  
    & \begin{tabular}{cccc}0.889 & 0.926 & 0.838 & 0.036\end{tabular}\\
GPONet \cite{gponet} & PR24  & \begin{tabular}{cccc}0.871 & 0.908 & \raisebox{5\height}{\rule{2.26em}{0.4pt}} & 0.055\end{tabular} & \begin{tabular}{cccc}\textcolor{red}{0.942} & \textcolor{blue}{0.964} & \raisebox{5\height}{\rule{2.26em}{0.4pt}} & \textcolor{red}{0.021}\end{tabular} & \begin{tabular}{cccc}0.911 & \textcolor{red}{0.967} & \raisebox{5\height}{\rule{2.26em}{0.4pt}} & 0.023\end{tabular} 
& \begin{tabular}{cccc}\textcolor{red}{0.893} & \textcolor{red}{0.899} & \raisebox{5\height}{\rule{2.26em}{0.4pt}} & \textcolor{blue}{0.045}\end{tabular}  
    & \begin{tabular}{cccc}\textcolor{red}{0.919} & 0.936 & \raisebox{5\height}{\rule{2.26em}{0.4pt}} & \textcolor{blue}{0.027}\end{tabular}\\
MDFANet \cite{mdfanet} & NeuCom25  & \begin{tabular}{cccc}0.867 & 0.898 & \textcolor{blue}{0.830} & 0.062\end{tabular} & \begin{tabular}{cccc}0.928 & 0.952 & 0.915 & 0.034\end{tabular} & \begin{tabular}{cccc}0.922 & \textcolor{blue}{0.959} & \textcolor{blue}{0.903} & 0.028\end{tabular} 
& \begin{tabular}{cccc}0.849 & 0.888 & \textcolor{blue}{0.764} & 0.055\end{tabular}  
    & \begin{tabular}{cccc}0.897 & 0.928 & \textcolor{blue}{0.849} & 0.034\end{tabular}\\
 \hline
ARNet & 2025 & \begin{tabular}{cccc}\textcolor{red}{0.878} & \textcolor{red}{0.917} & \textcolor{red}{0.847} & \textcolor{red}{0.052}\end{tabular} & \begin{tabular}{cccc}\textcolor{blue}{0.939} & \textcolor{red}{0.965} & \textcolor{red}{0.925} & \textcolor{blue}{0.024}\end{tabular} & \begin{tabular}{cccc}\textcolor{blue}{0.930} & 0.963 & \textcolor{red}{0.919} & \textcolor{blue}{0.024}\end{tabular} & \begin{tabular}{cccc}\textcolor{blue}{0.869} & \textcolor{blue}{0.896} & \textcolor{red}{0.807} & \textcolor{red}{0.044}\end{tabular} & \begin{tabular}{cccc}\textcolor{blue}{0.913} & \textcolor{red}{0.947} & \textcolor{red}{0.883} & \textcolor{red}{0.026}\end{tabular} \\ \hline
\end{tabular}
}
}
\end{table*}
\begin{table*}[!t]
\centering
\caption{Quantitative comparison of model complexity for reproducible or open-source COD methods in terms of the number of parameters (Params) and computational cost (FLOPs).
}
\setlength{\tabcolsep}{4pt}
\renewcommand{\arraystretch}{1.1} 
\label{tab:complexity}
\centering
{\fontsize{7}{8}\selectfont
\begin{tabular}{l|ccccccccccccc}
\hline
Method & SINet-v2 & C2FNet-v2 & ZoomNet & BGNet & FEDER & FSPNet & OAFormer & SARNet & DINet & Camoformer & PRNet & RISNet &ARNet \\ 
\hline
Params$\downarrow$ & 26.9M & 44.9M & 32.4M & 79.9M & 42.1M & 273.8M & 39.9M & 47.2M & 30.8M & 71.3M & 12.7M & 26.6M & 33.9M \\ 
FLOPs$\downarrow$  & 14.7G & 23.2G & 101.8G & 116.9G & 30.6G & 283.3G & --- & 23.1G & 112.2G & 47.1G & 10.2G & 39.1G & 54.1G \\ 
\hline
\end{tabular}}
\end{table*}
\subsection{Multi-scale Enhancement Module}

To further enhance the fused features, we incorporate a simple multi-scale enhancement module, which expands the receptive field using multi-scale convolutions~\cite{c2fnet-v2} and provides richer features for the CIIM to guide low-level features. Specifically, we construct a three-branch network where convolutional blocks of different scales process features in each branch to yield multi-scale features. As shown in Fig.~\ref{fig_4}, the three branches sequentially consist of two convolutional units with $5\times5$ kernels, two convolutional units with $3\times3$ kernels, and one $3\times3$ dilated convolutional unit. This can be expressed as:
\begin{equation}
\begin{cases}f_i^1 =C_{5\times5}BR(C_{5\times5}BR(f_{i})) \\
f_i^2 =C_{3\times3}BR(C_{3\times3}BR(f_i)) \\
f_i^3=DilC_{3\times3}BR(f_i)\end{cases}
\end{equation}
where $DilC_{3\times3}(\cdot)$ denotes a $3\times3$ dilated convolution, and in this paper we set the dilation rate to 3.

Next, the multi-scale features obtained from the three-branch network are concatenated. A $3\times3$ convolutional block is used to fuse the features while reducing the number of channels. The features are then constrained by residual connections with $f_i$ to enhance useful features. Finally, the output result $F_i$ is obtained through activation function processing, and the features are fed into the next layer of CIIM. The above process can be summarized as follows:
\begin{equation}
    F_i=ReLU(f_i\oplus C_{3\times3}BR(Cat(f_i^1,f_i^2,f_i^3)))
\end{equation}

\subsection{Loss Function}

The ARNet training employs a multi-level hybrid supervision strategy, supervising both the three predicted features and the boundary map generated by the model. For the predicted features, a joint supervision approach using Binary Cross-Entropy (BCE)~\cite{bce} loss and Intersection over Union (IoU) loss~\cite{iou} is adopted. These two losses further enhance discriminative capability and enable joint optimization. For the boundary prediction map, Dice loss~\cite{BGnet} is applied as a constraint. The overall loss function for the model can be defined as:
\begin{equation}\mathcal{L}=\sum_{i=1}^{3}(\mathcal{L}_{BCE}(P^{i},G)+\mathcal{L}_{IoU}(P^{i},G))+\lambda \mathcal{L}_{dice}(B,G)\end{equation}
with $\lambda=3$ to balance the boundary supervision, $\mathcal{L}_{BCE}$ denotes BCE loss, $\mathcal{L}_{IoU}$ represents IoU loss, $\mathcal{L}_{dics}$ denotes dice loss, $P$ denotes the predicted feature map, $B$ denotes the predicted boundary map, and $G$ denotes the ground truth feature map.

\section{Experimental Results and Analysis}

\begin{figure*}[!t]
\centering
\includegraphics[width=\textwidth]{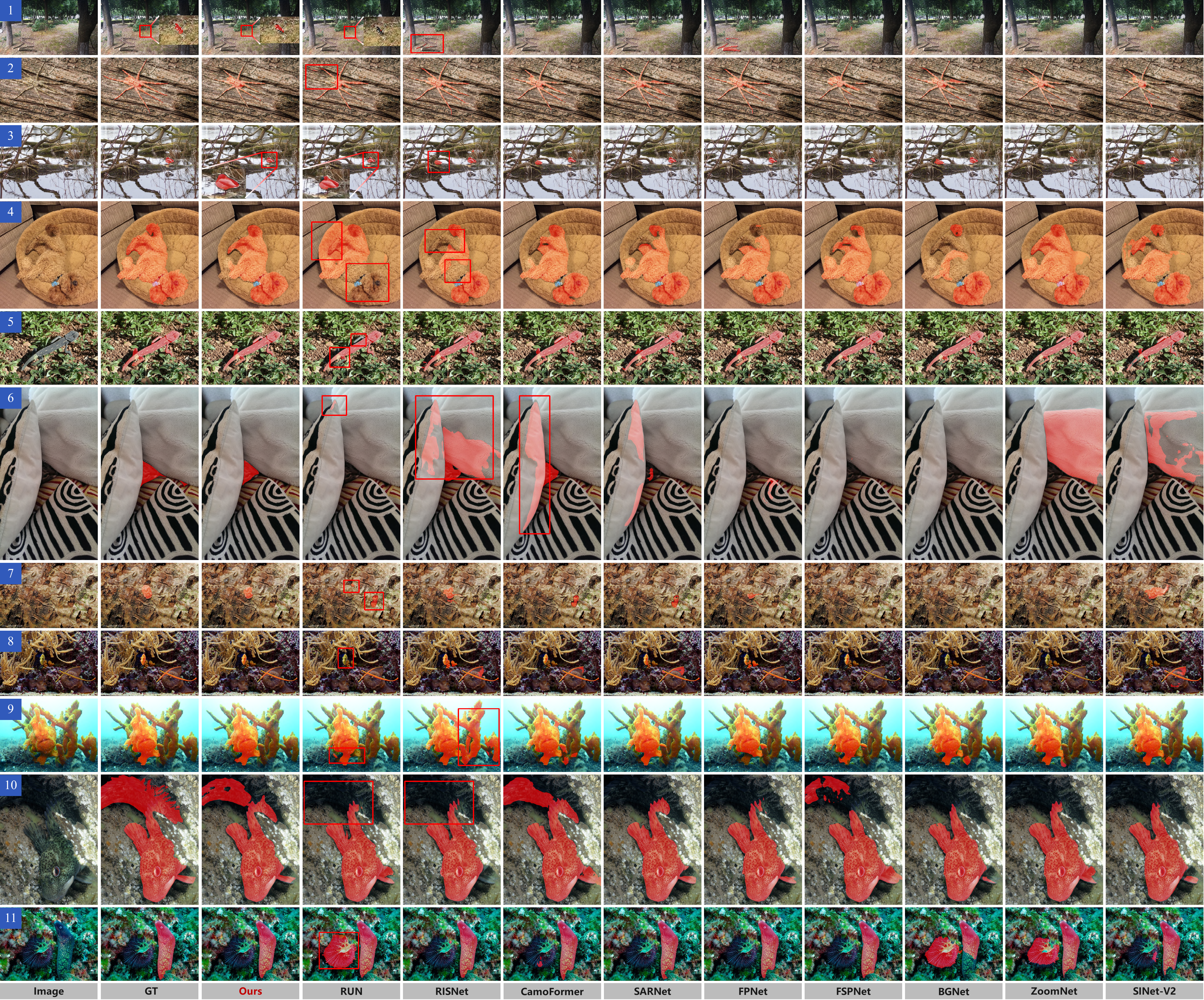}
\caption{Visual comparisons between the proposed ARNet and nine SOTA COD methods, where the red regions denote the predicted results. The selected scenes cover various challenging cases, including tiny objects (1st row), small objects (3rd and 7th rows), large objects (9th and 10th rows), multiple objects (8th row), multi-tentacle objects (2nd row), blurred boundaries (4th, 7th, and 10th rows), and occluded objects (6th row).}
\label{fig_5}
\end{figure*}
\subsection{Experimental Setups}

\textbf{Datasets.} For COD, four widely used benchmark datasets were employed: COD10K~\cite{sinet}, CAMO~\cite{camo}, NC4K~\cite{nc4k}, and CHAMELEON~\cite{chameleon}. COD10K comprises 3,040 training images and 2,026 test images, making it the largest benchmark dataset. CAMO contains 1,250 COD images, with 1,000 for training and 250 for testing. NC4K and CHAMELEON include 4,121 and 76 test images respectively, which are used to evaluate the model's generalization ability. For SOD, experiments are based on five existing mainstream datasets: DUT-OMRON~\cite{dut-om} (5,168), DUTS~\cite{duts} (10,553 + 5,017), ECSSD~\cite{ecssd} (1,000), HKU-IS~\cite{hku-is} (4,447), and PASCAL-S~\cite{pascal} (850). Following convention, training is performed using the 10,553 image training set from DUTS, while inference during the testing phase utilizes the remaining data.

\textbf{Evaluation Metrics.} For COD and SOD tasks, we selected four widely used metrics for evaluation: the S-measure ($S_{\alpha}$)~\cite{s} for assessing structural similarity, the E-measure ($E_{\phi}$)~\cite{e} for evaluating local and global pixel similarity, the F-measure ($F_{\beta}^{\omega}$)~\cite{f} for calculating the harmonic mean between precision and recall, and the Mean Absolute Error ($\mathcal{M}$)~\cite{m} for measuring the accuracy between predicted and ground-truth image pixels.

\textbf{Implementation Details.} The model is implemented using the PyTorch framework and trained on an NVIDIA 4090 GPU. During training, bilinear interpolation is employed to resize input images to a resolution of $416 \times 416$. The Adam optimizer~\cite{adam} was used to update parameters, with an initial learning rate of 5e-5. The learning rate decayed by a factor of 10 every 100 epochs. The batch size was set to 8, and the maximum number of epochs was set to 150.

\subsection{Comparison with SOTA Methods}

\textbf{COD Task Evaluation.} As shown in Table~\ref{tab1}, we quantitatively compared our proposed method with 17 other SOTA approaches, including SINet-v2~\cite{sinet-v2}, C2FNet-v2~\cite{c2fnet-v2}, ZoomNet~\cite{zoomnet}, BGNet~\cite{BGnet}, FEDER~\cite{feder}, FSPNet~\cite{fspnet}, OAFormer~\cite{oaformer}, FPNet~\cite{fpnet}, SARNet~\cite{SARNET}, DINet~\cite{dinet}, Camoformer~\cite{camoformer}, PRNet~\cite{prnet}, DRFNet~\cite{drfnet}, RISNet~\cite{wang2024depth}, CSFIN \cite{LI2025126451}, PRBE-Net~\cite{prbenet}, and RUN~\cite{run}. Our model achieves outstanding results across four commonly used evaluation metrics on four benchmark datasets, ranking first in 12 out of 16 comparisons and second in 4. Notably, by leveraging boundary priors and object localization maps for spatial auxiliary guidance, ARNet significantly enhances the structural similarity metric $S_{\alpha}$, achieving optimal performance across all four datasets with margins of 0.57\%, 1.37\%, 0.78\%, and 0.55\% over the method RUN. Table~\ref{tab:complexity} provides a quantitative comparison of model complexity for reproducible or open-source COD methods in terms of parameters and computational cost, offering a clear reference for efficiency analysis. Our method effectively controls model complexity while maintaining high accuracy, which can be attributed to the lightweight design of our modules, as detailed in Table~\ref{tab:params}. Fig.~\ref{fig_5} presents a qualitative analysis with visual comparisons against nine other representative methods. Eleven samples were selected, covering scenarios such as tiny objects, small objects, large objects, multiple objects, multi-tentacle objects and objects with a blurred boundary, demonstrating ARNet's robust generalization capability. Leveraging its dual-dimensional decoding framework and auxiliary information guidance, the model not only fully utilizes multi-level complementary information but also ensures consistent decoding semantics. Furthermore, the comparisons in rows 4, 7, and 10 of Fig.~\ref{fig_5} demonstrate the ability to accurately define boundaries in complex environments, proving the precision of ARNet's boundary extraction and guidance.

\begin{figure}
\centering
\includegraphics[width=\columnwidth]{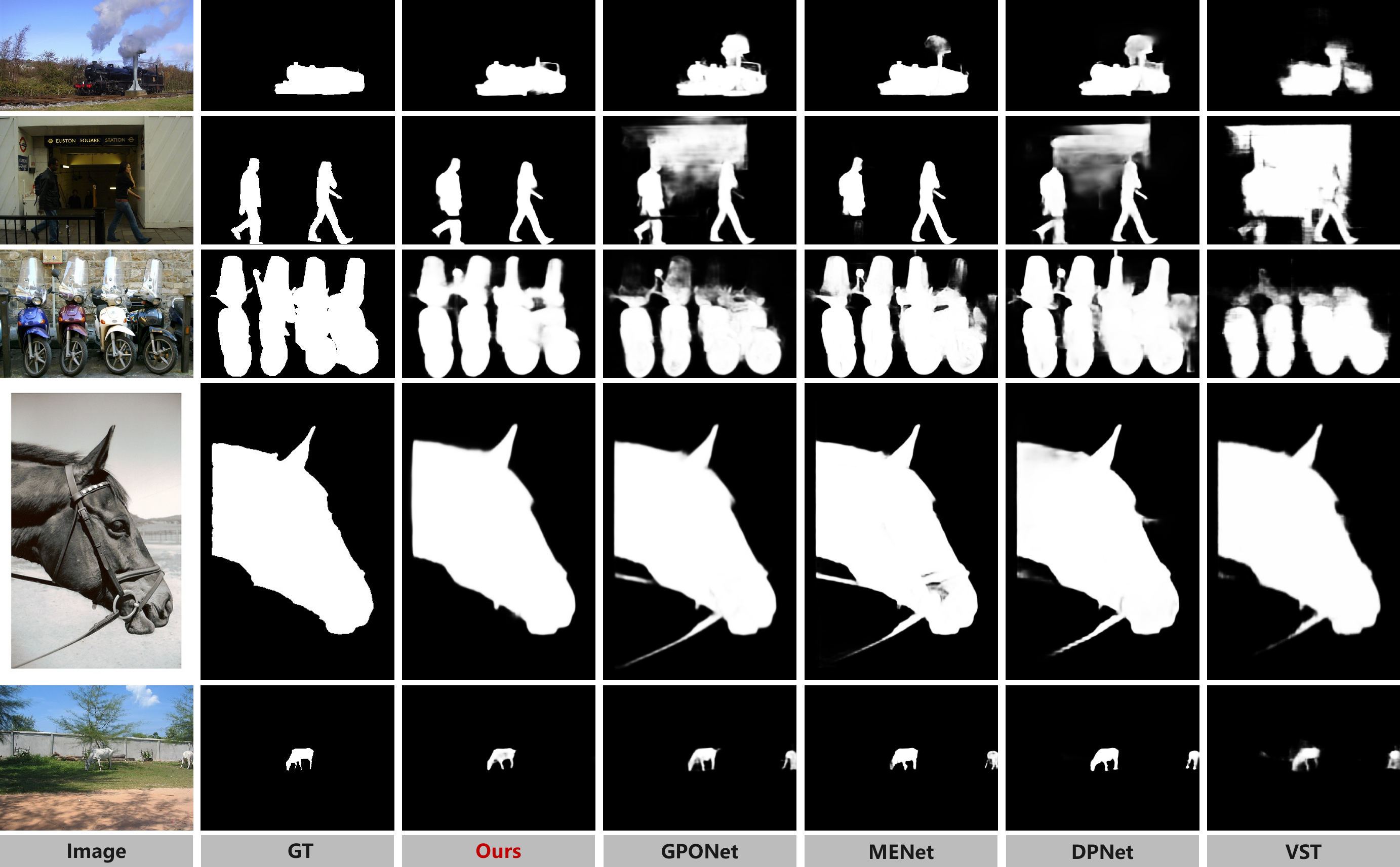}
\caption{Visual comparison of our proposed ARNet with four other SOTA SOD methods.}
\label{fig_6}
\end{figure}

\textbf{SOD Task Evaluation.} To validate the universality of ARNet's dual-dimensional decoding and auxiliary guidance, we extended our experiments to the similar segmentation task SOD. As shown in Table~\ref{tab2}, we compared nine SOTA methods across five mainstream SOD datasets: DNA~\cite{dna}, VST~\cite{vst}, CFNet~\cite{cfnet}, DPNet~\cite{dpnet}, ICON~\cite{icon}, MENet~\cite{menet}, MPTNet~\cite{mptnet}, GPONet~\cite{gponet}, and MDFANet~\cite{mdfanet}. Results demonstrate that ARNet outperforms other models overall, confirming the method's versatility. ARNet achieves superior performance across all metrics on the PASCAL-S dataset. Qualitative analysis in Fig.~\ref{fig_6} provides an intuitive comparison of ARNet's performance against other methods, highlighting ARNet's outstanding ability to localize salient objects and segment fine details.

\subsection{Ablation Studies}

\begin{figure*}[!t]
\centering
\includegraphics[width=\textwidth]{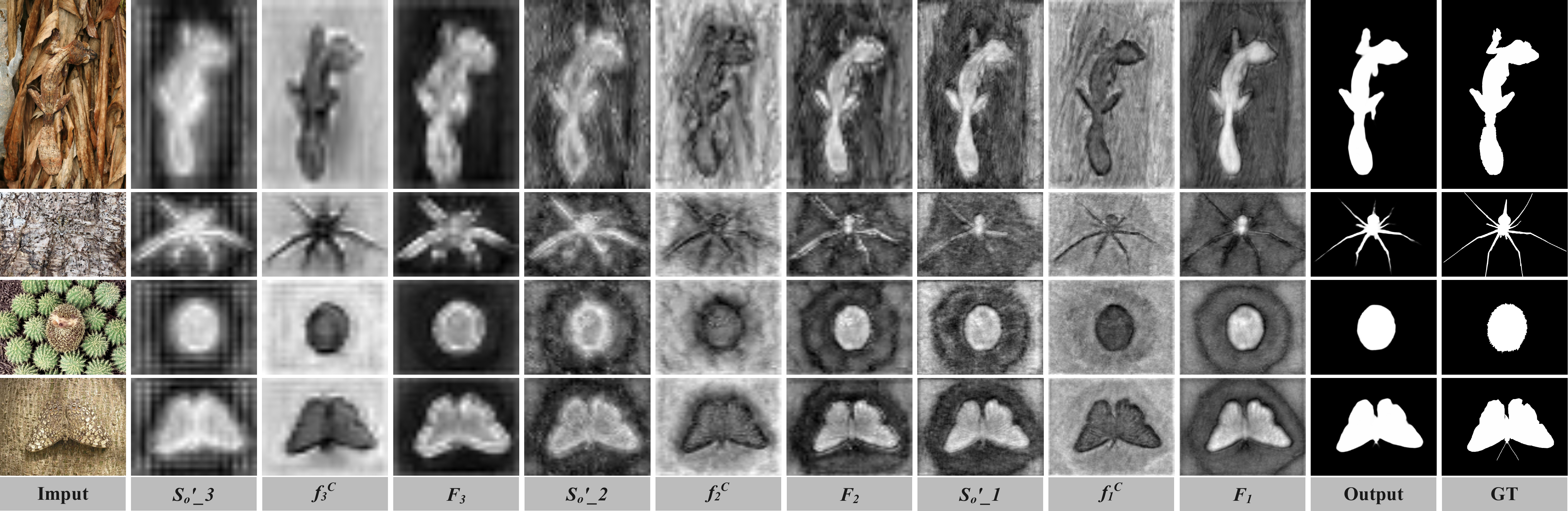}
\caption{Feature maps obtained at different positions in the ARNet network. During visualization, we average each feature along the channel dimension. From left to right, the outputs correspond to the various stages of refinement, from high-level to low-level features.}
\label{binary}
\end{figure*}

To validate the effectiveness of each module and component, we performed comprehensive ablation studies, covering the backbone, four modules, and two components, for a thorough evaluation. Table~\ref{tab3} details the ablation settings and experimental comparison data for the backbone and the four modules of our model. Table~\ref{tab:params} presents the ablation study of parameters and FLOPs for each component. Table~\ref{tab4} presents the ablation results for components FAP and HGA within the CIIM module. Table~\ref{attention} compares the performance of our FAP with other channel attention mechanisms. In each experiment, we removed or modified the ablation module within our ARNet standard network while keeping all other components unchanged. Moreover, we further study the feature maps obtained at different stages of the network, as illustrated in Fig.~\ref{binary}, to gain deeper insight into the effectiveness of each component.

\begin{table}[!t]
  \centering
  \caption{Ablation studies, with best results highlighted in \textbf{bold}.}
  \label{tab3}
  \resizebox{\linewidth}{!}{
  \begin{tabular}{l|l|cccccc} 
    \hline
    Datasets & Metrics  &Backbone & w/o CIIM & w/o MSE & w/o RE & w/o BE & ARNet \\
    \hline
    \multirow{4}{*}{COD10K-Test}   & $S_{\alpha} \uparrow$       & 0.848 & 0.876      & 0.877     & 0.879          & 0.878          & \textbf{0.883}          \\
                                 & $E_{\phi} \uparrow$         & 0.921 & 0.934 & 0.937  & 0.938         & \textbf{0.940}          & 0.938         \\
                                 & $F_{\beta}^{\omega} \uparrow$  & 0.732& 0.798          & \textbf{0.809} & 0.804          & 0.807          & 0.808 \\
                                 & $\mathcal{M}\downarrow$           &  0.028 & 0.022  & \textbf{0.021}         & 0.022          & \textbf{0.021}         & \textbf{0.021} \\
    \hline
    \multirow{4}{*}{CAMO-Test} & $S_{\alpha} \uparrow$       & 0.872 & 0.880          & 0.879          & 0.879       & 0.879    & \textbf{0.890} \\
                                 & $E_{\phi} \uparrow$         & 0.926 & 0.926   & 0.929        & 0.931          & 0.932          & \textbf{0.937} \\
                                 & $F_{\beta}^{\omega} \uparrow$  &0.814 & 0.836   & 0.844        & 0.838          & 0.840          & \textbf{0.853} \\
                                 & $\mathcal{M}\downarrow$         & 0.049& 0.045     & 0.044          & 0.044          & 0.044          & \textbf{0.042} \\
    \hline
    \multirow{4}{*}{NC4K}        & $S_{\alpha} \uparrow$       & 0.878 & 0.895 & 0.891  & 0.897         & 0.893 & \textbf{0.899}          \\
                                 & $E_{\phi} \uparrow$        & 0.931  & 0.939 & 0.938  & 0.941         & 0.940 & \textbf{0.942}          \\
                                 & $F_{\beta}^{\omega} \uparrow$ &0.813 &  0.851 & 0.854  & 0.854         & 0.851 & \textbf{0.857}          \\
                                 & $\mathcal{M}\downarrow$           & 0.036  & 0.030 & 0.031   & 0.030        & 0.031          & \textbf{0.029} \\
    \hline
    \multirow{4}{*}{CHAMELEON}   & $S_{\alpha} \uparrow$     &  0.891  & 0.907          & 0.906 & 0.914  & 0.910         & \textbf{0.921} \\
                                 & $E_{\phi} \uparrow$        &  0.944 & 0.953          & 0.956          & 0.962 & 0.961          & \textbf{0.964} \\
                                 & $F_{\beta}^{\omega} \uparrow$  &0.822 & 0.863          & 0.866          & 0.875& 0.872           & \textbf{0.883} \\
                                 & $\mathcal{M}\downarrow$          & 0.030  & 0.024          & 0.024          & 0.022 & 0.022           & \textbf{0.019} \\
    \hline
  \end{tabular}
  }
\end{table}

\begin{table}[!t]
  \centering
  \caption{Ablation study of component Params and FLOPs.} 
  \setlength{\tabcolsep}{4pt} 
  \renewcommand{\arraystretch}{1.1} 
  \label{tab:params} 
  \resizebox{\linewidth}{!}{
  \begin{tabular}{l|cccccccc}
    \hline
    Settings & Backbone & w/o CIIM & w/o HGA & w/o FAP & w/o MSE & w/o RE & w/o BE & ARNet \\
    \hline
    Params$\downarrow$ & 31.73M & 32.23M & 33.50M & 32.66M & 33.51M & 33.65M & 33.61M & 33.93M \\
    FLOPs$\downarrow$  & 28.33G & 37.26G & 52.10G & 38.77G & 48.07G & 51.45G & 51.90G & 54.14G \\
    \hline
  \end{tabular}}
\end{table}

\textbf{Effectiveness of CIIM Module and Components FAP, HGA.} The CIIM module comprises two key components: the channel-wise decoder and the hybrid guided attention mechanism, forming the core of our model. To thoroughly validate the effectiveness of the CIIM module, we conducted three sets of ablation experiments. As shown in Table~\ref{tab3}, we removed the entire CIIM from ARNet, instead fusing the input backbone features through simple addition and similarly combining the guidance maps via addition. As observed, removing CIIM resulted in decreased metrics across all four datasets, with a more pronounced decline compared to other modules. With CIIM incorporated, the $F_{\beta}^{\omega}$ improved by 1.25\%, 2.03\%, 0.71\%, and 2.32\% across the four datasets, respectively. As the harmonic mean of precision and recall, the overall increase in $F_{\beta}^{\omega}$ indicates that the CIIM module effectively enhances the model's balanced and robust segmentation performance across diverse datasets. 

To validate the effectiveness of individual CIIM components, two additional experiments were conducted (as shown in Table~\ref{tab4}), performing ablation tests on FAP and HGA separately. Comparisons with removing the entire CIIM module and the network as a whole demonstrated the components' contributions and complementarity. Notably, incorporating FAP yielded the largest improvement in $S_{\alpha}$, indicating that channel-dimensionality-assisted decoding effectively enhances object structure decoding. Further experiments comparing FAP with Squeeze-and-Excitation (SE~\cite{hu2018squeeze}) (as shown in Table~\ref{attention}) demonstrate that while SE shows some performance improvement, FAP outperforms SE overall, particularly in the $S_{\alpha}$ metric. This suggests that FAP’s unique ability to assist in channel-dimensionality decoding is more effective than SE’s reweighting mechanism, which primarily focuses on channel attention without the same level of structure-guided decoding.

\textbf{Effectiveness of RE and BE Modules.} In the RE and BE modules, we directly utilize backbone features to generate auxiliary guidance information for embedding into the HGA to guide feature decoding. Here, we remove them respectively to validate their effectiveness. The experimental results are shown in Table~\ref{tab3}. Overall, although the improvement in model performance from both modules is smaller than that from the MSE and CIIM modules, they still play an irreplaceable role and demonstrate good complementarity. Through the hybrid guidance of boundaries and regions, ARNet effectively suppresses pseudo-boundary interference and improves the accuracy of object spatial localization.

\begin{table*}[!t]
\caption{Ablation study of the FAP and HGA core components of CIIM, with best results highlighted in \textbf{bold}.}
\label{tab4}
\centering
\setlength{\tabcolsep}{5pt}
\renewcommand{\arraystretch}{1.0}
{\fontsize{7}{8.5}\selectfont
\resizebox{\linewidth}{!}{
\begin{tabular}{cc|cccc|cccc|cccc|cccc} 
\hline
\multirow{2}{*}{FAP} & \multirow{2}{*}{HGA} & \multicolumn{4}{c|}{COD10K-Test} & \multicolumn{4}{c|}{CAMO} & \multicolumn{4}{c|}{NC4K} & \multicolumn{4}{c}{CHAMELEON} \\ 

\cline{3-18}

 & &
$S_{\alpha} \uparrow$  & $E_{\phi} \uparrow$  &  $F_{\beta}^{\omega} \uparrow$ &  $\mathcal{M}\downarrow$  &
$S_{\alpha} \uparrow$  & $E_{\phi} \uparrow$  &  $F_{\beta}^{\omega} \uparrow$ &  $\mathcal{M}\downarrow$  &
$S_{\alpha} \uparrow$  & $E_{\phi} \uparrow$  &  $F_{\beta}^{\omega} \uparrow$ &  $\mathcal{M}\downarrow$  &
$S_{\alpha} \uparrow$  & $E_{\phi}\uparrow$  &  $F_{\beta}^{\omega} \uparrow$ &  $\mathcal{M}\downarrow$  \\ \hline

    &            & 0.876 & 0.934 & 0.798 & 0.022 & 0.880 & 0.926 & 0.836 & 0.045 & 0.895 & 0.939 & 0.851 & 0.030 & 0.907 & 0.953 & 0.863 & 0.024 \\
\checkmark &       & 0.880 & \textbf{0.939} & 0.810 & \textbf{0.021} & 0.877 & 0.928 & 0.837 & 0.045 & 0.896 & 0.941 & 0.856 & 0.030 & 0.918 & 0.962 & 0.881 & 0.020 \\
    & \checkmark & 0.878 & 0.938 & \textbf{0.811} & \textbf{0.021} & 0.879 & 0.933 & 0.846 & 0.043 & 0.894 & 0.940 & \textbf{0.858} & \textbf{0.029} & 0.912 & 0.961 & 0.880 & 0.021 \\
\checkmark & \checkmark & \textbf{0.883} & 0.938 & 0.808 & \textbf{0.021} & \textbf{0.890} & \textbf{0.937} & \textbf{0.853} & \textbf{0.042} & \textbf{0.899} & \textbf{0.942} & 0.857 & \textbf{0.029} & \textbf{0.921} & \textbf{0.964} & \textbf{0.883} & \textbf{0.019} \\ \hline
\end{tabular}
}
}
\end{table*}

\begin{table*}[!t]
\caption{Comparison of SE and FAP components on different datasets, with best results highlighted in \textbf{bold}.}
\label{attention}
\centering
\setlength{\tabcolsep}{5pt}
\renewcommand{\arraystretch}{1.0}
{\fontsize{7}{8.5}\selectfont
\resizebox{\linewidth}{!}{
\begin{tabular}{c|cccc|cccc|cccc|cccc} 
\hline
\multirow{2}{*}{Type} & \multicolumn{4}{c|}{COD10K-Test} & \multicolumn{4}{c|}{CAMO} & \multicolumn{4}{c|}{NC4K} & \multicolumn{4}{c}{CHAMELEON} \\ \cline{2-17}
 & $S_{\alpha} \uparrow$ & $E_{\phi} \uparrow$ & $F_{\beta}^{\omega} \uparrow$ & $\mathcal{M} \downarrow$ & $S_{\alpha} \uparrow$ & $E_{\phi} \uparrow$ & $F_{\beta}^{\omega} \uparrow$ & $\mathcal{M} \downarrow$ & $S_{\alpha} \uparrow$ & $E_{\phi} \uparrow$ & $F_{\beta}^{\omega} \uparrow$ & $\mathcal{M} \downarrow$ & $S_{\alpha} \uparrow$ & $E_{\phi} \uparrow$ & $F_{\beta}^{\omega} \uparrow$ & $\mathcal{M} \downarrow$ \\ \hline
SE & 0.881 & \textbf{0.940} & \textbf{0.813} & \textbf{0.020} & 0.884 & 0.934 & 0.850 & \textbf{0.042} & 0.896 & 0.941 & \textbf{0.859} & \textbf{0.029} & 0.915 & 0.961 & 0.878 & 0.022 \\
FAP & \textbf{0.883} & 0.938 & 0.808 & 0.021 & \textbf{0.890} & \textbf{0.937} & \textbf{0.853} & \textbf{0.042} & \textbf{0.899} & \textbf{0.942} & 0.857 & \textbf{0.029} & \textbf{0.921} & \textbf{0.964} & \textbf{0.883} & \textbf{0.019} \\ \hline
\end{tabular}
}
}
\end{table*}

\begin{table}[!t]
\caption{Quantitative evaluation results of different polyp segmentation methods on five benchmark datasets. The best performing results are marked in \textcolor{red}{red}, while the second best results are highlighted in \textcolor{blue}{blue}.}
\label{tab5}
\centering
\setlength{\tabcolsep}{6pt}
\renewcommand{\arraystretch}{0.95}
{\fontsize{6}{8}\selectfont
\resizebox{\linewidth}{!}{
\begin{tabular}{l|ccccc}
\hline
\multirow{2}{*}{Method} & \multicolumn{5}{c}{CVC-300 \cite{cvc300}} \\ \cline{2-6}
 & mDice$\uparrow$ & mIoU$\uparrow$ & $S_{\alpha}\uparrow$ & $F_{\beta}^{\omega}\uparrow$ & $E_{\phi}^{max}\uparrow$ \\ \hline
SFA \cite{sfa}    & 0.467 & 0.329 & 0.640 & 0.341 & 0.817 \\
PraNet \cite{fan2020pranet} & 0.871 & 0.797 & 0.925 & 0.843 & 0.972 \\
SANet \cite{sanet}   & 0.888 & 0.815 & 0.928 & 0.859 & 0.972 \\
LDNet~\cite{ldnet}               & 0.850 & 0.781 & 0.907 & 0.822 & 0.949 \\
CFANet \cite{cfanet} & 0.893 & 0.827 & 0.938 & 0.875 & \textcolor{blue}{0.978} \\
NPDNet \cite{npdnet} & \textcolor{blue}{0.908} & \textcolor{blue}{0.844} & \textcolor{blue}{0.939} & \textcolor{blue}{0.880} & 0.977 \\
ARNet               & \textcolor{red}{0.913} & \textcolor{red}{0.851} & \textcolor{red}{0.947} & \textcolor{red}{0.902} & \textcolor{red}{0.983} \\ \hline
\end{tabular}
}
}

{\fontsize{6}{8}\selectfont
\resizebox{\linewidth}{!}{
\begin{tabular}{l|ccccc}
\hline
\multirow{2}{*}{Method} & \multicolumn{5}{c}{CVC-ColonDB \cite{colondb}} \\ \cline{2-6}
 & mDice$\uparrow$ & mIoU$\uparrow$ & $S_{\alpha}\uparrow$ & $F_{\beta}^{\omega}\uparrow$ & $E_{\phi}^{max}\uparrow$ \\ \hline
SFA \cite{sfa}    & 0.469 & 0.347 & 0.634 & 0.379 & 0.765 \\
PraNet \cite{fan2020pranet} & 0.709 & 0.640 & 0.819 & 0.696 & 0.869 \\
SANet \cite{sanet}   & 0.753 & 0.670 & \textcolor{blue}{0.837} & 0.726 & 0.878 \\
LDNet~\cite{ldnet}              & 0.752 & 0.678 & 0.834 & 0.732 & 0.867 \\
CFANet \cite{cfanet} & 0.743 & 0.665 & 0.835 & 0.728 & \textcolor{red}{0.898} \\
NPDNet \cite{npdnet} & \textcolor{blue}{0.764} & \textcolor{blue}{0.682} & 0.835 & \textcolor{blue}{0.734} & 0.887 \\
ARNet               & \textcolor{red}{0.781} & \textcolor{red}{0.707} & \textcolor{red}{0.855} & \textcolor{red}{0.763} & \textcolor{blue}{0.889} \\ \hline
\end{tabular}
}
}

{\fontsize{6}{8}\selectfont
\resizebox{\linewidth}{!}{
\begin{tabular}{l|ccccc}
\hline
\multirow{2}{*}{Method} & \multicolumn{5}{c}{CVC-ClinicDB \cite{clinicdb}} \\ \cline{2-6}
 & mDice$\uparrow$ & mIoU$\uparrow$ & $S_{\alpha}\uparrow$ & $F_{\beta}^{\omega}\uparrow$ & $E_{\phi}^{max}\uparrow$ \\ \hline
SFA \cite{sfa}    & 0.700 & 0.607 & 0.793 & 0.647 & 0.885 \\
PraNet \cite{fan2020pranet} & 0.899 & 0.849 & 0.936 & 0.896 & 0.979 \\
SANet \cite{sanet}   & 0.916 & 0.859 & 0.939 & 0.909 & 0.976 \\
LDNet~\cite{ldnet}              & 0.909 & 0.856 & 0.933 & 0.898 & 0.972 \\
CFANet \cite{cfanet} & \textcolor{red}{0.933} & \textcolor{red}{0.883} & \textcolor{red}{0.950} & \textcolor{red}{0.924} & \textcolor{red}{0.989} \\
NPDNet \cite{npdnet} & \textcolor{blue}{0.925} & \textcolor{blue}{0.876} & \textcolor{blue}{0.947} & \textcolor{blue}{0.917} & 0.976 \\
ARNet               & 0.919 & 0.868 & 0.941 & 0.915 & \textcolor{blue}{0.978} \\ \hline
\end{tabular}
}
}

{\fontsize{6}{8}\selectfont
\resizebox{\linewidth}{!}{
\begin{tabular}{l|ccccc}
\hline
\multirow{2}{*}{Method} & \multicolumn{5}{c}{Kvasir \cite{kvasir}} \\ \cline{2-6}
 & mDice$\uparrow$ & mIoU$\uparrow$ & $S_{\alpha}\uparrow$ & $F_{\beta}^{\omega}\uparrow$ & $E_{\phi}^{max}\uparrow$ \\ \hline
SFA \cite{sfa}    & 0.723 & 0.611 & 0.782 & 0.670 & 0.849 \\
PraNet \cite{fan2020pranet} & 0.898 & 0.840 & 0.915 & 0.885 & 0.948 \\
SANet \cite{sanet}   & 0.904 & 0.847 & 0.915 & 0.892 & 0.953 \\
LDNet~\cite{ldnet}               & 0.902 & 0.847 & 0.912 & 0.889 & 0.945 \\
CFANet \cite{cfanet} & \textcolor{red}{0.915} & \textcolor{red}{0.861} & \textcolor{red}{0.924} & \textcolor{red}{0.903} & \textcolor{red}{0.962} \\
NPDNet \cite{npdnet} & 0.905 & 0.850 & 0.914 & 0.889 & 0.950 \\
ARNet               & \textcolor{blue}{0.910} & \textcolor{blue}{0.856} & \textcolor{blue}{0.923} & \textcolor{blue}{0.902} & \textcolor{blue}{0.958} \\ \hline
\end{tabular}
}
}

{\fontsize{6}{8}\selectfont
\resizebox{\linewidth}{!}{
\begin{tabular}{l|ccccc}
\hline
\multirow{2}{*}{Method} & \multicolumn{5}{c}{ETIS \cite{etis}} \\ \cline{2-6}
 &mDice$\uparrow$ & mIoU$\uparrow$ & $S_{\alpha}\uparrow$ & $F_{\beta}^{\omega}\uparrow$ & $E_{\phi}^{max}\uparrow$ \\ \hline
SFA \cite{sfa}    & 0.297 & 0.217 & 0.557 & 0.231 & 0.633 \\
PraNet \cite{fan2020pranet} & 0.628 & 0.567 & 0.794 & 0.600 & 0.841 \\
SANet \cite{sanet}   & \textcolor{blue}{0.750} & 0.654 & \textcolor{blue}{0.849} & 0.685 & \textcolor{blue}{0.897} \\
LDNet~\cite{ldnet}               & 0.605 & 0.542 & 0.754 & 0.565 & 0.774 \\
CFANet \cite{cfanet} & 0.732 & 0.655 & 0.845 & \textcolor{blue}{0.693} & 0.892 \\
NPDNet \cite{npdnet} & 0.737 & \textcolor{blue}{0.659} & 0.848 & \textcolor{blue}{0.693} & 0.891 \\
ARNet               & \textcolor{red}{0.784} & \textcolor{red}{0.711} & \textcolor{red}{0.871} & \textcolor{red}{0.749} & \textcolor{red}{0.916} \\ \hline
\end{tabular}
}
}
\end{table}

\textbf{Effectiveness of the MSE Module.} The MSE module is designed to expand the receptive field of the object and enrich object features through a multi-scale strategy. To validate the effectiveness of the MSE module, we removed three MSE modules from the network and directly supervised the features output by the CIIM, feeding them into the next layer of CIIM. As shown in Table~\ref{tab3}, removing the MSE modules resulted in a significant performance decline. For instance, on the CHAMELEON dataset, the top three metrics decreased by 1.63\%, 0.83\%, and 1.93\% respectively, while $M$ increased by 26.3\%. This is because in complex environments like COD, multi-scale information plays a crucial role in object recognition.

\textbf{Feature Map Visualization Study.} To further evaluate and demonstrate the working mechanism of our proposed method, we conduct a detailed analysis of the feature maps obtained at different stages of the network, as shown in Fig.~\ref{binary}. We select the ${S_{o}^{\prime}}_{-}3$, $f_3^C$, and $F_3$ from each layer, which represent the feature maps processed by the channel dimension decoder, mixed attention guidance, and MSE module, respectively. As can be seen, overall, the components of the network progressively refine the features from high-level to low-level, with each component working in harmony while fulfilling its specific role. Specifically, using ${S_{o}^{\prime}}_{-}3$, $f_3^C$, and $F_3$ as examples, the backbone features, after passing through our channel dimension decoder, generate rough object regions and contours. After boundary priors and region localization guidance, $f_3^C$ is obtained, with a significant improvement in object clarity and sharper boundaries. Finally, after processing with the MSE module, $F_3$ is obtained, which enhances texture information and enriches feature representations to guide and supplement the subsequent processing stages.

\subsection{Application to Polyp Segmentation}

Due to the high similarity between polyps and normal colonic mucosa, accurate segmentation and resection of polyps are crucial for the diagnosis of colorectal cancer. Given the similarity between COD and polyp segmentation tasks, we extend ARNet to polyp segmentation to validate its effectiveness in downstream applications. In our experiments, we utilized five commonly used public datasets: CVC-300~\cite{cvc300}, CVC-ColonDB~\cite{colondb}, CVC-ClinicDB~\cite{clinicdb}, Kvasir~\cite{kvasir}, and ETIS~\cite{etis}. Following the established experimental protocol, we trained our model on the combined set of 900 images from Kvasir and 550 images from CVC-ClinicDB, with the remaining data from all datasets reserved for testing. For evaluation, we employed five widely used metrics: mDice~\cite{colondb}, mIoU~\cite{colondb}, $S_{\alpha}$~\cite{s}, $F_{\beta}^{\omega}$~\cite{f}, and $E_{\phi}^{max}$~\cite{e}.
\begin{figure}
\centering
\includegraphics[width=\columnwidth]{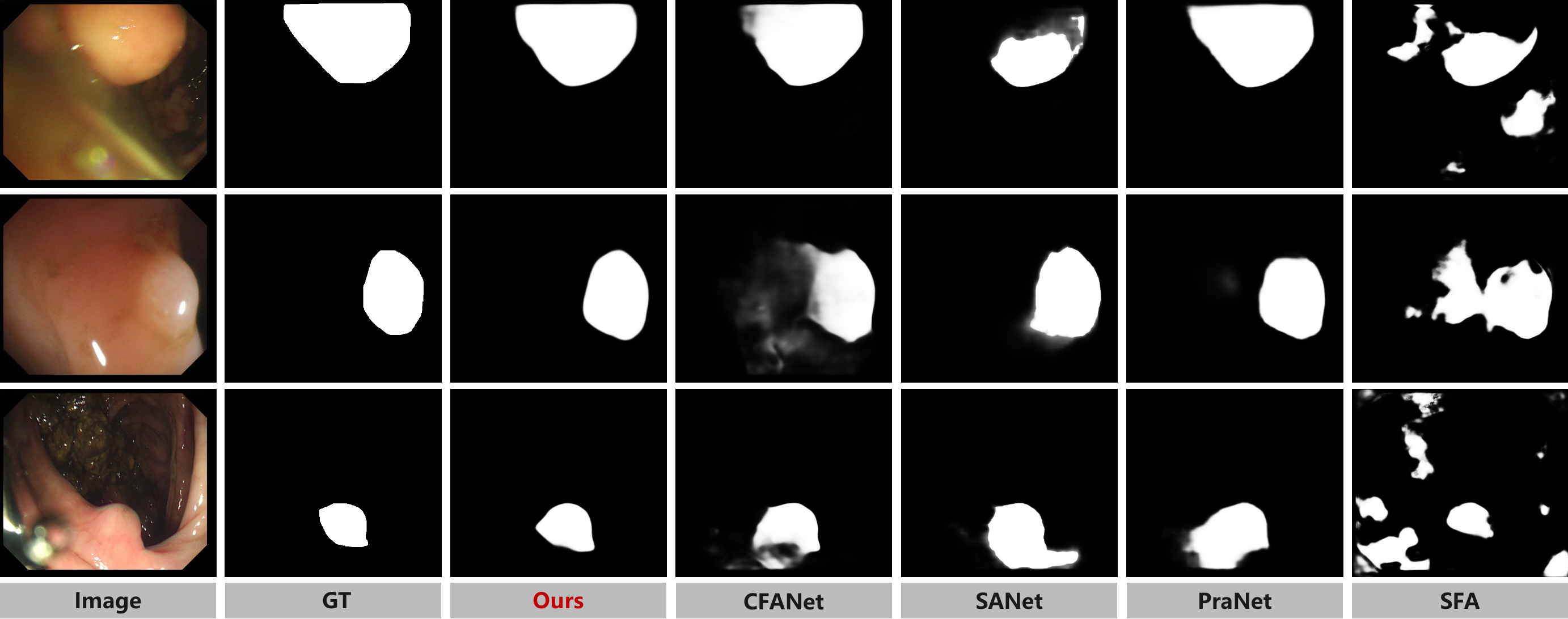}
\caption{Qualitative evaluation results of different polyp segmentation methods.}
\label{fig_7}
\end{figure}

\textbf{Quantitative Evaluation.} We quantitatively compare ARNet with SFA~\cite{sfa}, PraNet~\cite{fan2020pranet}, SANet~\cite{sanet}, LDNet~\cite{ldnet}, CFANet~\cite{cfanet}, and NPDNet~\cite{npdnet}. The experimental results are shown in Table~\ref{tab5}, where ARNet achieves outstanding performance. Particularly on the ETIS dataset, compared to NPDNet, ARNet achieves improvements of 6.38\%, 7.89\%, 2.71\%, 8.08\%, and 2.81\% in mDice, mIoU, $S_{\alpha}$, $F_{\beta}^{\omega}$, and $E_{\phi}^{max}$, respectively.

\textbf{Qualitative Evaluation.} Fig.~\ref{fig_7} presents a visual comparison between ARNet and four representative networks, featuring three challenging samples. Compared to other methods, ARNet demonstrates superior segmentation capabilities due to the powerful representational power of its bidimensional hybrid decoder, outperforming alternatives in both polyp boundary detection and spatial localization.

\subsection{Application to Transparent Object Detection}

Transparent objects such as all-glass facades and windows often impede the operation of robots and drones, posing potential safety hazards. However, conventional assistive technologies rarely address the detection of such safety-critical objects. For instance, robots performing tasks in living rooms or offices must avoid transparent, fragile items like glass and vases; similarly, drones require their visual navigation systems to identify glass walls and windows during flight to prevent collisions.

\begin{figure}
\centering
\includegraphics[width=\columnwidth]{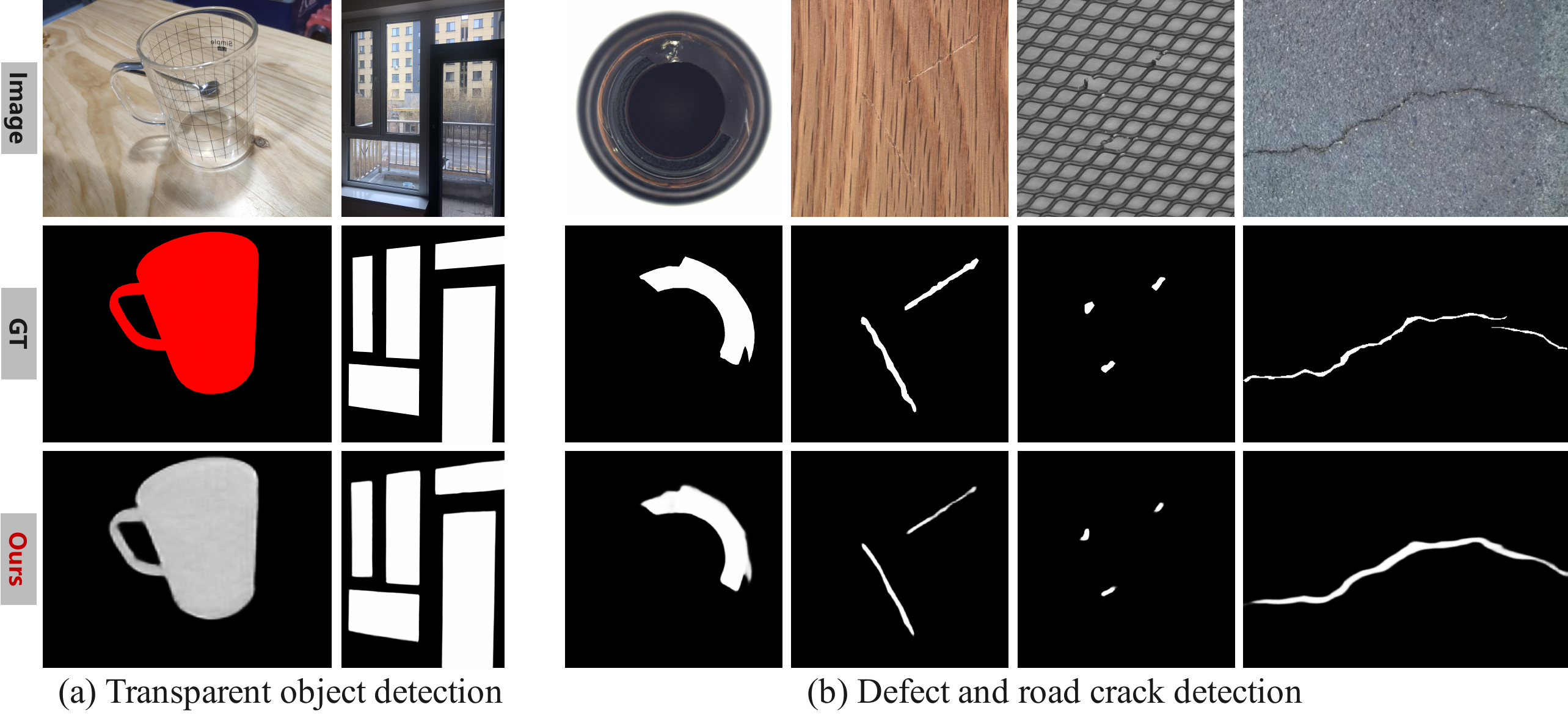}
\caption{Visual results for transparent object detection, defect detection, and road crack detection. The first to third rows show the original image, ground truth (GT), and ARNet segmentation results, respectively.}
\label{fig_8}
\end{figure}

To validate the effectiveness of this method for transparent object segmentation, we conducted experimental evaluations on the Trans10K dataset~\cite{trans10k}. This dataset contains images from diverse scenes including living rooms, offices, supermarkets, kitchens, and tables, categorizing objects into two main groups: large, fixed transparent objects like walls and windows, and movable objects such as bottles. The primary objective of this experiment is to enhance performance in transparent object segmentation and demonstrate the model's applicability to this task. For consistency and convenience, the experimental setup aligns with the COD experiments, utilizing the training set from Trans10K. Visualization results, as shown in Fig.~\ref{fig_8}(a), validate the effectiveness of our approach in transparent object detection.

\subsection{Application to Defect and Road Crack Detection}

In scenarios such as industrial defect detection and road crack detection, the issue of low-contrast objects against backgrounds is equally prominent, presenting challenges common to both these tasks and the COD task. Defective products frequently occur in industrial production, while road crack detection finds extensive application in scenarios like drone-based road inspections. Accurate detection in these contexts is crucial for minimizing economic losses. In this experiment, we utilize the CSD2K~\cite{fan2023advances} dataset, which consolidates five defect segmentation databases into a mixed dataset comprising 2,492 samples. Fig.~\ref{fig_8}(b) displays the segmentation results of our model across four scenarios: bottles, wooden boards, grid patterns, and road cracks. Compared to ground truth, our ARNet accurately identifies and segments objects, demonstrating the potential of our approach for downstream tasks.

\section{Conclusion}
Considering two key limitations of existing COD methods: 1) Insufficient cross-channel information interaction within same-layer features; 2) Lack of effective collaborative modeling between boundary and region information. To address these issues, this paper proposes a novel ARNet. This design incorporates a dual-dimensional decoding architecture, integrating CIIM, BE, and RE modules to perform hybrid decoding across both layer and channel dimensions while providing precise boundary and region guidance. This approach effectively captures complementary information between feature channels, enables joint learning of object regions and boundaries, and further enriches contextual representations through the MSE module. Extensive experiments demonstrate that ARNet achieves SOTA performance across multiple COD and SOD benchmark datasets, while its robust generalization capability is further validated on salient object detection and various downstream applications. Beyond establishing a new performance benchmark for COD, ARNet also provides a general and powerful paradigm for addressing broader visual challenges involving low-contrast or low-distinctiveness segmentation. This work lays a solid foundation for future research into more structured and explicit feature interaction mechanisms in deep learning models.

\section*{CRediT authorship contribution statement}
\textbf{Kuan Wang}: Conceptualization, Methodology, Software, Validation, Formal analysis, Visualization, Writing -- original draft. \textbf{Yanjun Qin}: Methodology, Validation, Writing -- review \& editing.
\textbf{Mengge Lu}: Software, Data curation, Visualization.
\textbf{Liejun Wang}: Resources, Supervision, Writing -- review \& editing.
\textbf{Xiaoming Tao}: Conceptualization, Supervision, Project administration, Funding acquisition, Writing -- review \& editing.

\section*{Declaration of competing interest}
The authors declare that they have no known competing financial interests or personal relationships
that could have appeared to influence the work reported in this paper.

\section*{Acknowledgments}
This work was supported in part by the Xinjiang Cheonchi Talent Introduction Program, in part by the Xinjiang Efficient Scientific Research Business Funds Project under Grant XJEDU2025J001, and in part by the Xinjiang Graduate Innovation and Practice Program under Grant XJ2025G087.

\section*{Data availability}
The implementation code and trained models of the proposed ARNet are available at: \url{https://github.com/akuan1234/ARNet-v2}.

\bibliographystyle{cas-model2-names}
\bibliography{bib}

\end{document}